\documentclass[sigconf]{acmart}

\renewcommand\footnotetextcopyrightpermission[1]{} 
\usepackage{tabularx}
\usepackage{bbding}
\usepackage{multirow}
\usepackage{color}
\usepackage{subcaption}
\usepackage[justification=centering]{caption}
\usepackage{arydshln}
\usepackage{adjustbox}

\AtBeginDocument{%
}

\setcopyright{acmcopyright}
\copyrightyear{2023}
\acmYear{2023}
\acmDOI{XXXXXXX.XXXXXXX}

\acmConference[arXiv]
{}{July, 2023}{Nanjing, China}

\acmSubmissionID{2678}

\begin{document}

\title{MaxSR: Image Super-Resolution Using Improved MaxViT}

\author{Bincheng Yang}
\email{yangbincheng@hotmail.com}
\orcid{0000-0002-3903-5425}
\author{Gangshan Wu}
\email{gswu@nju.edu.cn}
\orcid{0000-0003-1391-1762}
\affiliation{%
\institution{State Key Laboratory for Novel Software Technology, Nanjing University}
\streetaddress{163 Xianlin Avenue, Qixia District}
\city{Nanjing}
\state{Jiangsu}
\country{P.R.China}
\postcode{210023}
}

\begin{abstract}
While transformer models have been
demonstrated to be
effective for natural language processing tasks and high-level vision tasks,
only a few attempts have been made to use powerful transformer models for single image super-resolution.
Because transformer models have powerful representation capacity and the in-built self-attention mechanisms in transformer models
help to
leverage self-similarity prior in input low-resolution image to improve performance for single image super-resolution,
we present a single image super-resolution model based on recent hybrid vision transformer of MaxViT, named as MaxSR.
MaxSR consists of four parts, a shallow feature extraction block, multiple cascaded adaptive MaxViT blocks to extract deep hierarchical features and model global self-similarity from low-level features efficiently, a hierarchical feature fusion block, and finally a reconstruction block.
The key component of MaxSR, i.e., adaptive MaxViT block, is based on MaxViT block which mixes MBConv with squeeze-and-excitation,
block attention and grid attention.
In order to achieve better global modelling of self-similarity in input low-resolution image, we improve
block attention and grid attention
in MaxViT block to adaptive
block attention and adaptive grid attention
which do self-attention inside each window across all grids and each grid across all windows respectively in the most efficient way.
We instantiate proposed model for classical single image super-resolution (MaxSR) and lightweight single image super-resolution (MaxSR-light).
Experiments show that our MaxSR and MaxSR-light establish new state-of-the-art performance efficiently.
\end{abstract}



\maketitle

\section{Introduction}
\begin{figure}[t]
\begin{center}
\centering
\begin{minipage}{.96\linewidth}
\centering
\includegraphics[width=\linewidth]{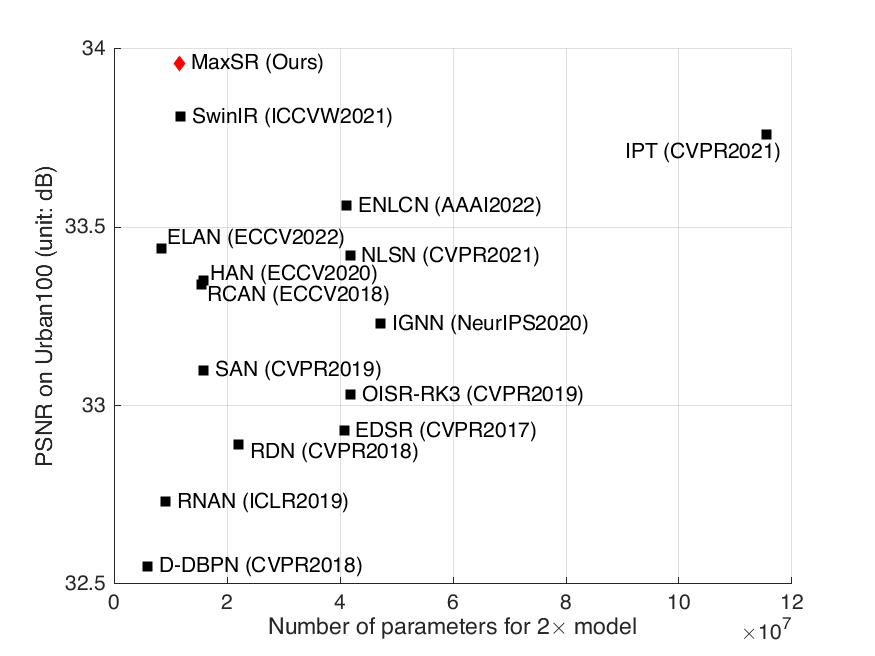}
\end{minipage}
\end{center}
\caption{Comparison for Performance and Number of Parameters of Different Methods.
}
\label{fig:param_psnr}
\end {figure}
\label{sec:intro}

Single image super-resolution (SISR) is an ill-posed inverse problem because the input low-resolution (LR) image can be mapped to many high-resolution (HR) undegraded images with different details. It has a lot of applications, such as security and surveillance
\cite{DBLP:journals/tip/ZouY12},
medical image
\cite{DBLP:conf/miccai/ShiCLZBBMDOR13}
and satellite and aerial imaging \cite{liebel2016single}.
How to use different image priors such as
self-similarity prior \cite{DBLP:books/daglib/0070474,DBLP:conf/cvpr/ZontakI11} is the key to tackle challenging ill-posed SISR problem.

Deep convolutional neural networks (CNNs) have been successful to learn a direct mapping between input LR image and output HR image for SISR in recent years \cite{dong2016accelerating,shi2016real,johnson2016perceptual,kim2016accurate,DBLP:conf/cvpr/KimLL16,DBLP:conf/cvpr/TaiY017,lai2017deep,lim2017enhanced,haris2018deep,DBLP:conf/iccv/QiuWT019,hui2018fast,li2018multi,mao2016image,ledig2017photo,tong2017image,tai2017memnet,zhang2018residual,DBLP:conf/nips/LiuWFLH18,DBLP:conf/iclr/ZhangLLZF19,DBLP:conf/cvpr/DaiCZXZ19,DBLP:conf/cvpr/MeiFZHHS20,DBLP:conf/nips/ZhouZZL20,mei2021image,DBLP:conf/aaai/XiaHTYLZ22}.
However, while transformer models have been shown to be
effective
for natural language processing tasks and high-level vision tasks,  
only a few attempts have been made to use transformer models for SISR \cite{DBLP:conf/cvpr/Chen000DLMX0021,DBLP:conf/iccvw/LiangCSZGT21,DBLP:conf/eccv/ZhangZGZ22}.

Due to their content-dependent filtering ability and effective long-range dependency modelling ability in contrast to deep CNNs,
transformer models have powerful representation capacity
which can be used to model complex non-linear mappings between input LR image and output HR image.
Moreover,
the inbuilt self-attention (SA) mechanisms in transformer models
help to
leverage self-similarity prior \cite{DBLP:books/daglib/0070474,DBLP:conf/cvpr/ZontakI11}
in input LR image to
improve
performance for SISR.
Nevertheless, because the feature maps in SISR network models have
very big spatial sizes and the computational complexity of self-attention is quadratic order with respect to the number of attention locations,
there is a dilemma between long-range dependency modelling ability/self-similarity modelling ability and efficiency.
Therefore, how to utilizing powerful representation capacity and self-similarity modelling ability of transformer models effectively in an efficiency way to improve performance for SISR is an interesting challenge.
\begin{figure*}[t]
\begin{center}
\centering
\begin{minipage}{.96\linewidth}
\centering
\includegraphics[width=\linewidth]{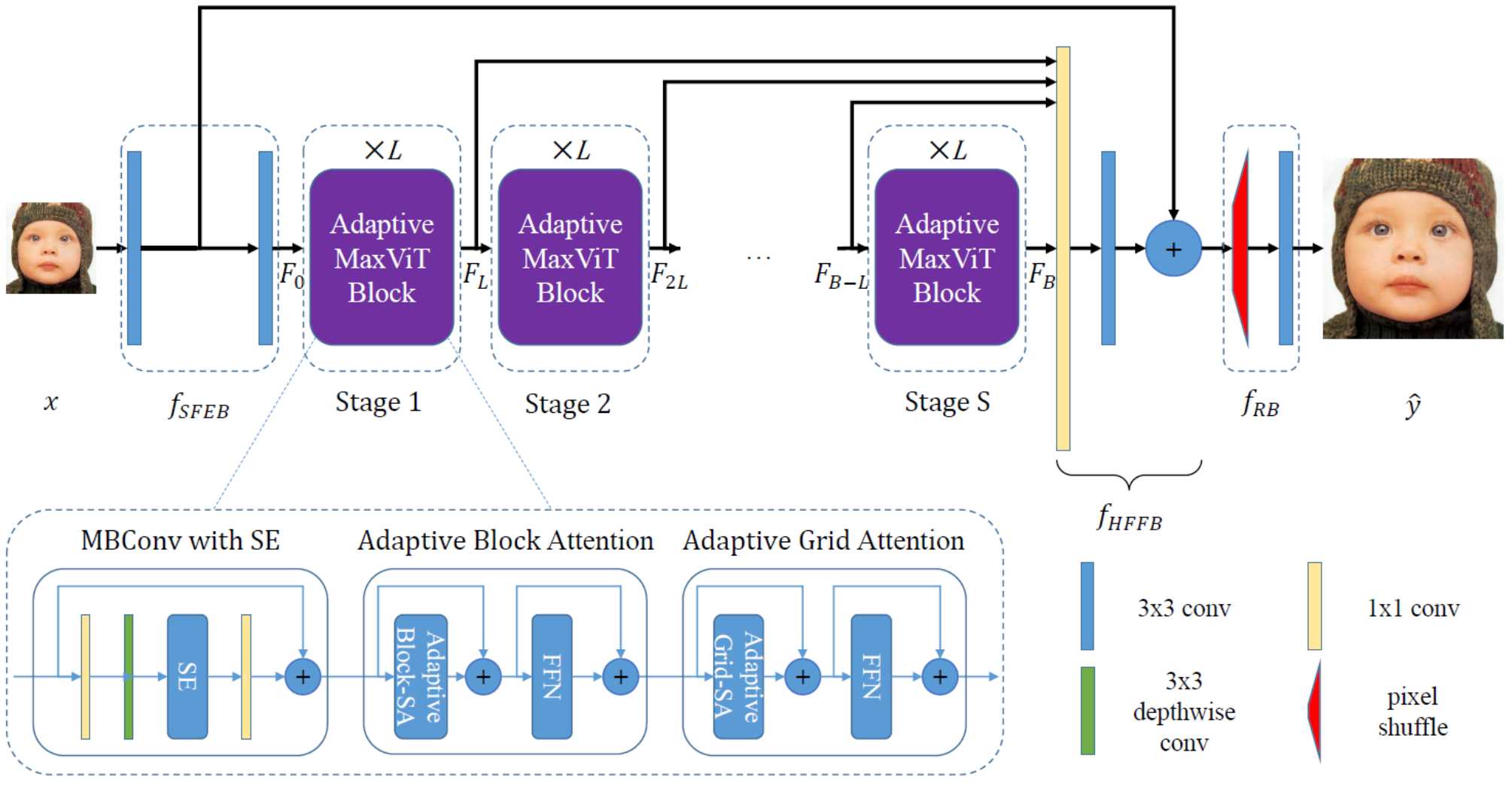}
\end{minipage}
\end{center}
\caption{The Architecture of MaxSR.
}
\label{fig:arch_MaxSR}
\end{figure*}

In order to
utilize powerful representation capacity and self-similarity modelling ability of transformer models for SISR effectively and efficiently,
we present a novel single image super-resolution
model based on recent hybrid transformer of MaxViT \cite{DBLP:conf/eccv/TuTZYMBL22},
named as MaxSR
which achieves state-of-the-art performance efficiently as shown in Figure \ref{fig:param_psnr}.
MaxSR is shown in Figure \ref{fig:arch_MaxSR}, which consists
of four parts, a shallow feature extraction block (SFEB), multiple cascaded adaptive MaxViT blocks,
a hierarchical feature fusion block
(HFFB), and finally a reconstruction blocks (RB).
The SFEB uses convolution layers to extract low-level features.
The key component
of MaxViT,
adaptive MaxViT block,
is based on MaxViT block which mixes MBConv with squeeze-and-excitation (SE), block attention and grid attention.
In order to achieve better global modelling of self-similarity in input
LR image, we improve block attention and grid attention in MaxViT block to
adaptive block attention and adaptive grid attention
to form adaptive MaxViT block,
which
has
optimal sub-quadratic complexity
with respect to the spatial size of input feature map
to integrate information from all windows for each grid and integrate information from all grids for each window.
Multiple cascaded adaptive MaxViT blocks are used to extract deep hierarchical features and model global self-similarity in input image
in an efficient way.
The HFFB fuses hierarchical feature maps in the model for reconstruction of output HR image.
The RB uses pixel shuffle layers and a convolutional layer to reconstruct output HR image.

In summary, our main contributions are as follows:
\begin{itemize}
\item We present a novel single image super-resolution
model based on recent hybrid vision transformer of MaxViT
for single image super-resolution to utilize powerful
representation capacity and self-similarity modelling ability of transformer models to improve performance for SISR efficiently.
\item We improve
MaxViT block to
adaptive MaxViT block in order to model self-similarity globally in input LR image in a better way, which can integrate information from all windows for each grid
and integrate information from all grids for each window
with an optimal sub-quadratic complexity with respect to the spatial size of input feature map to be efficient and scalable.
\item Ablation study demonstrates effectiveness of adaptiveness of MaxViT block, relative position embedding, and adaptive block attention or adaptive grid attention for SISR.
\item Experiments show our model establishes new state-of-the-art performance for classical and lightweight single image super-resolution tasks efficiently.
\end{itemize}

\section{Related Work}
\label{sec:blind}
Since the pioneer work of Dong \MakeLowercase{\textit{et al.}} \cite{dong2016image} uses a three layer convolutional neural network (SRCNN)
to learn a direct mapping from an input LR image to an output HR image for SISR,
a lot of deep
CNN based methods \cite{dong2016accelerating,shi2016real,johnson2016perceptual,kim2016accurate,DBLP:conf/cvpr/KimLL16,DBLP:conf/cvpr/TaiY017,lai2017deep,lim2017enhanced,haris2018deep,DBLP:conf/iccv/QiuWT019,hui2018fast,li2018multi,mao2016image,ledig2017photo,tong2017image,tai2017memnet,zhang2018residual} have been proposed to boost performance and efficiency for SISR by utilizing various network design strategies, especially
residual connections and dense connections.

Attention mechanisms are used by deep CNN based methods to focus on more important information to improve performance. Channel attention mechanism was first introduced into a
CNN (RCAN) by Zhang \MakeLowercase{\textit{et al.}} \cite{zhang2018image} to solve SISR. Dai \MakeLowercase{\textit{et al.}} design a second-order attention network (SAN) \cite{DBLP:conf/cvpr/DaiCZXZ19} to utilize second-order information to compute attention scores for more powerful feature correlation and feature expression learning, which also utilizes non-local attention to model long-range dependencies. To dynamically select appropriate kernel size to adjust the size of receipt field, Zhang \MakeLowercase{\textit{et al.}} design a kernel attention network \cite{zhang2020kernel} for SISR. A holistic attention network (HAN) was proposed by Niu \MakeLowercase{\textit{et al.}} \cite{niu2020single} to model holistic interdependencies between layers, channels and positions.

The re-occurrences of small patches in the same image are demonstrated as a strong image prior in natural images \cite{DBLP:books/daglib/0070474,DBLP:conf/cvpr/ZontakI11}, which can be utilized by non-local attention to improve reconstruction performance for SISR \cite{DBLP:conf/nips/LiuWFLH18,DBLP:conf/iclr/ZhangLLZF19,DBLP:conf/cvpr/DaiCZXZ19,DBLP:conf/cvpr/MeiFZHHS20,DBLP:conf/nips/ZhouZZL20,mei2021image,DBLP:conf/aaai/XiaHTYLZ22}.
Non-local operations were first introduced by Liu \MakeLowercase{\textit{et al.}} \cite{DBLP:conf/nips/LiuWFLH18} into a recurrent neural network (NLRN) to improve parameter efficiency and capture feature correlation for image restoration. To utilize local and non-local attention blocks to attend to challenging parts and capture long-range dependencies, Zhang \MakeLowercase{\textit{et al.}} \cite{DBLP:conf/iclr/ZhangLLZF19} propose residual non-local attention network (RNAN). To utilize cross-scale non-local self-similarity of patches in input image for SISR, Mei \MakeLowercase{\textit{et al.}} \cite{DBLP:conf/cvpr/MeiFZHHS20} propose a cross-scale non-local attention network (CSNLN), Zhou \MakeLowercase{\textit{et al.}} \cite{DBLP:conf/nips/ZhouZZL20} propose a cross-scale internal graph neural network (IGNN).
To boost reconstruction performance and efficiency, a deep network (NLSN) combining non-local operations and sparse representation was proposed by Mei \MakeLowercase{\textit{et al.}} \cite{mei2021image}.
Xia \MakeLowercase{\textit{et al.}} \cite{DBLP:conf/aaai/XiaHTYLZ22} propose an efficient non-local contrastive attention for image super-resolution (ENLCN) to improve performance for SISR by adopting kernel method to approximate exponential function efficiently and contrastive learning to
separate relevant and irrelevant features.

Because of content-based filtering ability and long-range dependency modelling ability of transformer models in contrast to
CNNs,
they have been used to achieve impressive performance for natural language processing tasks 
and high-level vision tasks
.
Besides the powerful representation capacity of transformer models, the inbuilt self-attention mechanisms of transformer models help to leverage self-similarity prior in nature images to improve performance for SISR.
A pre-trained image processing transformer (IPT)
based on transformer model
was proposed by Chen \MakeLowercase{\textit{et al.}} \cite{DBLP:conf/cvpr/Chen000DLMX0021} to jointly process different kinds of low-level vision tasks in one framework. Liang \MakeLowercase{\textit{et al.}} propose image restoration using swin transformer (SwinIR) \cite{DBLP:conf/iccvw/LiangCSZGT21} based on swin transformer which utilizes
self-attention with shifted windows to enhance long-range dependency modelling ability.
Zhang \MakeLowercase{\textit{et al.}} \cite{DBLP:conf/eccv/ZhangZGZ22} propose an efficient long-range attention network for image super-resolution (ELAN), which utilizes shift convolution and group-wise multi-scale self-attention to improve efficiency and performance for SISR.

\section{Method}
\subsection{Network Architecture}
The proposed MaxSR, as shown in Figure \ref{fig:arch_MaxSR}, consists of four parts: a shallow feature extraction block (SFEB),
multiple cascaded adaptive MaxViT blocks
(AMTBs),
a hierarchical feature fusion block (HFFB), and finally a reconstruction block (RB).
Let's denote \(x\) as the LR input and \(\hat{y}\) as the HR output of the network.

First, we use two $3\times3$ convolution layers in SFEB to extract low-level features from network input 
\(x\),
\begin{flalign}
& F_{-1} = Conv_3(x), \\
& F_{0} = f_{SFEB}(x) = Conv_3(F_{-1}), 
\end{flalign}
where \(f_{SFEB}\) denotes the function of SFEB, \(F_{0}\) denotes the extracted features to be sent to the first adaptive MaxViT block.

Second, we use multiple cascaded adaptive MaxViT blocks (AMTBs) to learn deep hierarchical features and model global self-similarity from low-level features efficiently. Suppose \(B\) blocks are stacked, we have
\begin{equation}
\begin{split}
F_{b} = f_{AMTB,b}(f_{AMTB,b-1}(...(f_{AMTB,1}(F_{0}))...)),
\end{split}
\end{equation}
where
$b\in\{1,2,...,B\}$,
\(f_{AMTB,b}\) denotes the function of \(b\)-th AMTB, \(F_{b}\) denote the output feature maps of the \(b\)-th AMTB.

Afterwards, we fuse hierarchical feature maps extracted by the AMTBs and SFEB using HFFB.
In order to fuse hierarchical feature maps extracted by the AMTBs efficiently,
we divide cascaded $B$ AMTBs into sequential $S$ stages evenly so each stage will have $B//S$ AMTBs (we ensure $B$ will be divided without remainder by $S$ and let $L=B//S$).
Then we fuse feature maps output by last AMTB of each stage,
so we only fuse feature maps every $L$ AMTBs,
\begin{equation}
\begin{split}
{H} = f_{HFFB}(F_{-1}, F_{L}, F_{2L}, ..., F_{B}),
\end{split}
\end{equation}
\begin{equation}
\begin{split}
{H} = Conv_3(Conv_1(Concat(F_{L}, F_{2L}, ..., F_{B})))+F_{-1},
\end{split}
\end{equation}
where \(F_{-1}\) denotes the output of first convolution layer of SFEB, \(f_{HFFB}\) denotes the function of HFFB and \(H\) denotes the output of HFFB.

Finally, we use fused features output by HFFB as input to RB to reconstruct the HR image \(y\),
\begin{equation}\hat{y} = g(x) = f_{RB}({H}) = Conv_3(PS(H)),\end{equation}
where $PS$ denotes the function of pixel shuffle layers, \(f_{RB}\) denotes the function of the RB, and \(g\) denotes the function of MaxSR.

Given a set of training image pairs \(\{x^{(m)}, y^{(m)}\}^{M}_{m=1}\), the network is used to minimize the following Mean Absolute Error (MAE) loss:
\begin{equation}L(\Theta)=\frac{1}{M}\sum^{M}_{m=1}||y^{(m)}-\hat{y}^{(m)}||_{1},\end{equation}
where \(\Theta\) is the parameters of MaxSR, which includes parameters in SFEB, AMTBs, HFFB and RB.
\subsection{Adaptive MaxViT Block}
Adaptive MaxViT block is shown in bottom-left of Figure \ref{fig:arch_MaxSR}, which is based on MaxViT block proposed in MaxViT \cite{DBLP:conf/eccv/TuTZYMBL22}. MaxViT block consists of a MB-Conv block \cite{DBLP:journals/corr/HowardZCKWWAA17} with squeeze-and-excitation (SE) module \cite{DBLP:conf/cvpr/HuSS18}, a block attention (block-SA)
with fixed attention footage such as $8\times8$ and a grid attention (grid-SA)
with fixed attention footage such as $8\times8$ sequentially.
The combination of block attention and grid attention sequentially is called Max-SA in MaxViT \cite{DBLP:conf/eccv/TuTZYMBL22}.

Because Max-SA adopts fixed size of attention locations for block attention and grid attention, it can not integrate information from all windows of block attention for each grid of grid attention and can not integrate information from all grids of grid attention for each window of block attention when the spatial size of input feature map is large enough, which will impair global modelling ability of self-similarity of Max-SA degrading performance for SISR.

In order to integrate information from all windows of block attention
for each grid of grid attention and integrate information from all grids of grid attention
for each window of block attention, we need
an fixed-size block attention
and an adaptive-size grid attention
with respect to the size of input feature map,
e.g., a block attention with a fixed footage of $8\times8$ and a grid attention with an adaptive size of $\lceil{H/8}\rceil\times\lceil{H/8}\rceil$ after padding the input feature map,
or
an adaptive-size block attention
and a fixed-size grid attention
with respect to the size of input feature map,
e.g., a block attention with an adaptive size of $\lceil{H/8}\rceil\times\lceil{H/8}\rceil$ and a grid attention with a fixed footage of $8\times8$ after padding the input feature map,
or
an adaptive-size block attention
and an adaptive-size grid attention
with respect to the size of input feature map,
e.g., a block attention with an adaptive size of
$\lceil{H^{\frac{1}{3}}}\rceil\times\lceil{W^{\frac{1}{3}}}\rceil$
and a grid attention with an adaptive size of
$\lceil{H^{\frac{1}{3}}}\rceil^2\times\lceil{W^{\frac{1}{3}}}\rceil^2$
after padding the input feature map.
If we use an adaptive-size block attention
and a fixed-size grid attention,
or an fixed-size block attention and an adaptive-size grid attention, the complexity of Max-SA will be quadratic order with respect to the spatial size of input feature map to be inefficient. So the choice will be utilizing an adaptive-size block attention and adaptive-size grid attention for Max-SA and balancing the complexity of two kinds of self-attention to achieve an optimal sub-quadratic complexity.
In order to integrate information from all windows of block attention for each grid of grid attention and integrate information from all grids of grid attention for each window of block attention meanwhile achieve the lowest complexity, the attention footage of block attention and grid attention need all to be approximately $\lceil\sqrt{H}\rceil\times\lceil\sqrt{W}\rceil$ adaptive to the size of input feature map.
Based on this analysis,
we propose to utilize 
adaptive Max-SA in
adaptive Max-ViT block
to improve performance for SISR, which adopts
balanced adaptive size of attention locations for block attention and grid attention to improve global modelling ability of Max-SA in the most efficient way
inspired by the strategy proposed in HiT \cite{DBLP:conf/nips/ZhaoZCMZ21}.

More specifically, given an input feature map $F \in \mathbb{R}^{H\times W\times C}$,
adaptive Max-SA first pads the input feature map into
a tensor of shape $[{\lceil\sqrt{H}\rceil}^2,{\lceil\sqrt{W}\rceil}^2, C]$
by zeros (padding the input feature map into a tensor of shape $[{\lceil\sqrt{H}\rceil}\times\lceil{H/\lceil\sqrt{H}\rceil}\rceil, {\lceil\sqrt{W}\rceil}\times\lceil{W/\lceil\sqrt{W}\rceil}\rceil, C]$ will be an efficient approximation),
and divides the padded input feature map into non-overlapping windows of
spatial size of
\(\lceil\sqrt{H}\rceil\times\lceil\sqrt{W}\rceil\).
Then adaptive Max-SA performs adaptive block attention within each window of spatial size of $\lceil\sqrt{H}\rceil\times\lceil\sqrt{W}\rceil$ and adaptive grid attention within each uniform grid of spatial size of $\lceil\sqrt{H}\rceil\times\lceil\sqrt{W}\rceil$ (performing grid attention within each uniform grid of spatial size of $\lceil{H/\lceil\sqrt{H}\rceil}\rceil\times\lceil{W/\lceil\sqrt{W}\rceil}\rceil$ will be an efficient approximation corresponding to above approximate padding).
In this way, adaptive Max-SA can integrate information from all windows of adaptive block attention for each grid
of adaptive grid attention and integrate information from all uniform grids of adaptive grid attention for each window
of adaptive block attention to improve information propagation and global modelling ability of similarity which is beneficial to performance for SISR.
The complexity of adaptive Max-SA will be
optimal sub-quadratic order of $O[(\lceil\sqrt{H}\rceil\times\lceil\sqrt{W}\rceil)^2\times(\lceil\sqrt{H}\rceil\times\lceil\sqrt{W}\rceil)]$
for our motivation.

Compared to original fixed-size Max-SA proposed in MaxViT \cite{DBLP:conf/eccv/TuTZYMBL22}, which can not integrate information from all windows of fixed size of block attention for each grid of grid attention and can not integrate information from all uniform grids of fixed size of grid attention for each window of block attention when the input image is large enough, adaptive Max-SA has better global modelling ability of self-similarity to improve performance for SISR.
Compared to shifted window based self-attention in SwinIR \cite{DBLP:conf/iccvw/LiangCSZGT21}, which can only integrate information from neighbor windows in each execution of self-attention after shifting windows, adaptive Max-SA has
more direct and faster information propagation between different windows or grids
which is helpful to global modelling of self-similarity
to improve performance for SISR.
\begin{table*}[tb]
{
\caption{
Ablation Study for Effect of Adaptive Block Attention or Adaptive Grid Attention.
}
\label{tab:ablation_study_block_or_grid}
\scriptsize
\begin{center}
{
\begin{tabularx}{.96\linewidth}{|X|c|c|c|c|c|c|c|c|c|c|c|c|}
\hline
\multirow{2}*{Method}&\multirow{2}*{Scale}&\multirow{2}*{Training Set}&\multicolumn{2}{|c|}{Set5}&\multicolumn{2}{|c|}{Set14}&\multicolumn{2}{|c|}{B100}&\multicolumn{2}{|c|}{Urban100}&\multicolumn{2}{|c|}{Manga109}\\
\cline{4-13}
&&&PSNR&SSIM&PSNR&SSIM&PSNR&SSIM&PSNR&SSIM&PSNR&SSIM\\
\hline
\hline
Ada-Block-Attention &$\times2$&DIV2K&38.28&0.9617&\textcolor{red}{34.31}&\textcolor{blue}{0.9242}&\textcolor{blue}{32.41}&0.9023&\textcolor{blue}{33.47}&0.9384&\textcolor{blue}{39.68}&\textcolor{blue}{0.9791}\\
Ada-Gird-Attention &$\times2$&DIV2K&\textcolor{blue}{38.31}&\textcolor{blue}{0.9618}&34.19&0.9235&\textcolor{blue}{32.41}&\textcolor{blue}{0.9026}&33.40&\textcolor{blue}{0.9391}&39.63&\textcolor{blue}{0.9791}\\
MaxSR &$\times2$&DIV2K&\textcolor{red}{38.35}&\textcolor{red}{0.9620}&\textcolor{blue}{34.20}&\textcolor{red}{0.9246}&\textcolor{red}{32.44}&\textcolor{red}{0.9027}&\textcolor{red}{33.59}&\textcolor{red}{0.9398}&\textcolor{red}{39.70}&\textcolor{red}{0.9792}\\
\hline
\end{tabularx}
\\
}
\end{center}
}
\end{table*}
\begin{table*}[tb]
{
\caption{
Ablation Study for Effect of Adaptiveness of MaxViT Block and Relative Position Embedding.
}
\label{tab:ablation_study_adaptiveness_and_relative_position_embedding}
\scriptsize
\begin{center}
{
\begin{tabularx}{.96\linewidth}{|X|c|c|c|c|c|c|c|c|c|c|c|c|}
\hline
\multirow{2}*{Method}&\multirow{2}*{Scale}&\multirow{2}*{Training Set}&\multicolumn{2}{|c|}{Set5}&\multicolumn{2}{|c|}{Set14}&\multicolumn{2}{|c|}{B100}&\multicolumn{2}{|c|}{Urban100}&\multicolumn{2}{|c|}{Manga109}\\
\cline{4-13}
&&&PSNR&SSIM&PSNR&SSIM&PSNR&SSIM&PSNR&SSIM&PSNR&SSIM\\
\hline
\hline
BP-Fix-SA-Tr-Fix-SA-Te &$\times2$&DIV2K&\textcolor{blue}{38.31}&\textcolor{blue}{0.9619}&34.09&0.9232&\textcolor{blue}{32.42}&\textcolor{blue}{0.9029}&33.40&0.9389&39.55&0.9786\\
BP-Fix-SA-Tr-Ada-SA-Te &$\times2$&DIV2K&38.30&0.9618&\textcolor{red}{34.25}&\textcolor{red}{0.9241}&\textcolor{blue}{32.42}&0.9027&33.50&0.9391&39.66&0.9786\\
BP-MaxSR &$\times2$&DIV2K&\textcolor{blue}{38.31}&\textcolor{blue}{0.9619}&34.18&\textcolor{blue}{0.9239}&32.39&\textcolor{red}{0.9032}&\textcolor{blue}{33.76}&\textcolor{blue}{0.9415}&\textcolor{blue}{39.76}&\textcolor{blue}{0.9792}\\
BP-MaxSR-RPE &$\times2$&DIV2K&\textcolor{red}{38.37}&\textcolor{red}{0.9621}&\textcolor{blue}{34.20}&0.9232&\textcolor{red}{32.45}&\textcolor{red}{0.9032}&\textcolor{red}{33.80}&\textcolor{red}{0.9417}&\textcolor{red}{39.80}&\textcolor{red}{0.9794}\\
\hline
\end{tabularx}
\\
}
\end{center}
}
\end{table*}
\begin{table*}[ht]
{
\caption{Average PSNR/SSIM Metrics for Scale Factors \(\times2\), \(\times3\), \(\times4\) and \(\times8\) of Compared Methods for Classical Image SR.
Best is in \textcolor{red}{Red}, and Second Best is in \textcolor{blue}{Blue}.
}
\label{tab:classical_pnsr_ssim_table}
\scriptsize
\begin{center}
{
\begin{tabularx}{.96\linewidth}{|X|c|c|c|c|c|c|c|c|c|c|c|c|}
\hline
\multirow{2}*{Method}&\multirow{2}*{Scale}&\multirow{2}*{Training Set}&\multicolumn{2}{|c|}{Set5}&\multicolumn{2}{|c|}{Set14}&\multicolumn{2}{|c|}{B100}&\multicolumn{2}{|c|}{Urban100}&\multicolumn{2}{|c|}{Manga109}\\
\cline{4-13}
&&&PSNR&SSIM&PSNR&SSIM&PSNR&SSIM&PSNR&SSIM&PSNR&SSIM\\
\hline
\hline
EDSR \cite{lim2017enhanced}&$\times2$&DIV2K&38.11&0.9602&33.92&0.9195&32.32&0.9013&32.93&0.9351&39.10&0.9773\\
RDN \cite{zhang2018residual}&$\times2$&DIV2K&38.24&0.9614&34.01&0.9212&32.34&0.9017&32.89&0.9353&39.18&0.9780\\
RCAN \cite{zhang2018image}&$\times2$&DIV2K&38.27&0.9614&34.12&0.9216&32.41&0.9027&33.34&0.9384&39.44&0.9786\\
NLRN \cite{DBLP:conf/nips/LiuWFLH18}&$\times2$&291&38.00&0.9603&33.46&0.9159&32.19&0.8992&31.81&0.9249&-&-\\
RNAN \cite{DBLP:conf/iclr/ZhangLLZF19}&$\times2$&DIV2K&38.17&0.9611&33.87&0.9207&32.32&0.9014&32.73&0.9340&39.23&0.9785\\
SAN \cite{DBLP:conf/cvpr/DaiCZXZ19}&$\times2$&DIV2K&38.31&\textcolor{blue}{0.9620}&34.07&0.9213&32.42&0.9028&33.10&0.9370&39.32&\textcolor{blue}{0.9792}\\
OISR-RK3 \cite{DBLP:conf/cvpr/HeMWLY019}&$\times2$&DIV2K&38.21&0.9612&33.94&0.9206&32.36&0.9019&33.03&0.9365&-&-\\
IGNN \cite{DBLP:conf/nips/ZhouZZL20}&$\times2$&DIV2K&38.24&0.9613&34.07&0.9217&32.41&0.9025&33.23&0.9383&39.35&0.9786\\
HAN \cite{niu2020single}&$\times2$&DIV2K&38.27&0.9614&34.16&0.9217&32.41&0.9027&33.35&0.9385&39.46&0.9785\\
NLSN \cite{mei2021image}&$\times2$&DIV2K&38.34&0.9618&34.08&0.9231&32.43&0.9027&33.42&0.9394&39.59&0.9789\\
SwinIR \cite{DBLP:conf/iccvw/LiangCSZGT21}&$\times2$&DIV2K&38.35&\textcolor{blue}{0.9620}&34.14&0.9227&32.44&\textcolor{blue}{0.9030}&33.40&0.9393&39.60&\textcolor{blue}{0.9792}\\
ENLCN \cite{DBLP:conf/aaai/XiaHTYLZ22}&$\times2$&DIV2K&\textcolor{blue}{38.37}&0.9618&34.17&0.9229&\textcolor{red}{32.49}&\textcolor{red}{0.9032}&33.56&\textcolor{blue}{0.9398}&39.64&0.9791\\
ELAN \cite{DBLP:conf/eccv/ZhangZGZ22}&$\times2$&DIV2K&38.36&\textcolor{blue}{0.9620}&\textcolor{blue}{34.20}&0.9228&32.45&\textcolor{blue}{0.9030}&33.44&0.9391&39.62&\textcolor{red}{0.9793}\\
MaxSR (Ours) &$\times2$&DIV2K&38.35&\textcolor{blue}{0.9620}&\textcolor{blue}{34.20}&\textcolor{blue}{0.9246}&32.44&0.9027&\textcolor{blue}{33.59}&\textcolor{blue}{0.9398}&\textcolor{blue}{39.70}&\textcolor{blue}{0.9792}\\
MaxSR+ (Ours) &$\times2$&DIV2K&\textcolor{red}{38.40}&\textcolor{red}{0.9621}&\textcolor{red}{34.34}&\textcolor{red}{0.9250}&\textcolor{blue}{32.46}&\textcolor{blue}{0.9030}&\textcolor{red}{33.71}&\textcolor{red}{0.9405}&\textcolor{red}{39.79}&\textcolor{blue}{0.9792}\\
\hdashline
D-DBPN \cite{haris2018deep}&$\times2$&DIV2K+Flickr2K&38.09&0.9600&33.85&0.9190&32.27&0.9000&32.55&0.9324&38.89&0.9775\\
IPT \cite{DBLP:conf/cvpr/Chen000DLMX0021}&$\times2$&ImageNet&38.37&-&34.43&-&32.48&-&33.76&-&-&-\\
SwinIR \cite{DBLP:conf/iccvw/LiangCSZGT21}&$\times2$&DIV2K+Flickr2K&38.42&0.9623&34.46&0.9250&\textcolor{blue}{32.53}&\textcolor{blue}{0.9041}&33.81&\textcolor{blue}{0.9427}&\textcolor{blue}{39.92}&0.9797\\
MaxSR (Ours) &$\times2$&DIV2K+Flickr2K&\textcolor{blue}{38.45}&\textcolor{blue}{0.9624}&\textcolor{blue}{34.54}&\textcolor{blue}{0.9258}&\textcolor{blue}{32.53}&0.9039&\textcolor{blue}{33.96}&0.9425&\textcolor{blue}{39.92}&\textcolor{blue}{0.9799}\\
MaxSR+ (Ours) &$\times2$&DIV2K+Flickr2K&\textcolor{red}{38.49}&\textcolor{red}{0.9625}&\textcolor{red}{34.63}&\textcolor{red}{0.9261}&\textcolor{red}{32.55}&\textcolor{red}{0.9042}&\textcolor{red}{34.10}&\textcolor{red}{0.9432}&\textcolor{red}{39.99}&\textcolor{red}{0.9800}\\
\hline
\hline
EDSR \cite{lim2017enhanced}&$\times3$&DIV2K&34.65&0.9280&30.52&0.8462&29.25&0.8093&28.80&0.8653&34.17&0.9476\\
RDN \cite{zhang2018residual}&$\times3$&DIV2K&34.71&0.9296&30.57&0.8468&29.26&0.8093&28.80&0.8653&34.13&0.9484\\
RCAN \cite{zhang2018image}&$\times3$&DIV2K&34.74&0.9299&30.65&0.8482&29.32&0.8111&29.09&0.8702&34.44&0.9499\\
NLRN \cite{DBLP:conf/nips/LiuWFLH18}&$\times3$&291&34.27&0.9266&30.16&0.8374&29.06&0.8026&27.93&0.8453&-&-\\
RNAN \cite{DBLP:conf/iclr/ZhangLLZF19}&$\times3$&DIV2K&34.66&0.9290&30.52&0.8462&29.26&0.8090&28.75&0.8646&34.25&0.9483\\
SAN \cite{DBLP:conf/cvpr/DaiCZXZ19}&$\times3$&DIV2K&34.75&0.9300&30.59&0.8476&29.33&0.8112&28.93&0.8671&34.30&0.9494\\
OISR-RK3 \cite{DBLP:conf/cvpr/HeMWLY019}&$\times3$&DIV2K&34.72&0.9297&30.57&0.8470&29.29&0.8103&28.95&0.8680&-&-\\
IGNN \cite{DBLP:conf/nips/ZhouZZL20}&$\times3$&DIV2K&34.72&0.9298&30.66&0.8484&29.31&0.8105&29.03&0.8696&34.39&0.9496\\
HAN \cite{niu2020single}&$\times3$&DIV2K&34.75&0.9299&30.67&0.8483&29.32&0.8110&29.10&0.8705&34.48&0.9500\\
NLSN \cite{mei2021image}&$\times3$&DIV2K&34.85&0.9306&30.70&0.8485&29.34&0.8117&29.25&0.8726&34.57&0.9508\\
SwinIR \cite{DBLP:conf/iccvw/LiangCSZGT21}&$\times3$&DIV2K&34.89&0.9312&30.77&0.8503&29.37&0.8124&29.29&0.8744&34.74&0.9518\\
ELAN \cite{DBLP:conf/eccv/ZhangZGZ22}&$\times3$&DIV2K&\textcolor{blue}{34.90}&\textcolor{blue}{0.9313}&\textcolor{blue}{30.80}&\textcolor{blue}{0.8504}&\textcolor{blue}{29.38}&0.8124&29.32&0.8745&34.73&0.9517\\
MaxSR (Ours) &$\times3$&DIV2K&34.84&0.9310&30.79&0.8503&\textcolor{blue}{29.38}&\textcolor{blue}{0.8126}&\textcolor{blue}{29.45}&\textcolor{blue}{0.8754}&\textcolor{blue}{34.92}&\textcolor{blue}{0.9522}\\
MaxSR+ (Ours) &$\times3$&DIV2K&\textcolor{red}{34.92}&\textcolor{red}{0.9315}&\textcolor{red}{30.87}&\textcolor{red}{0.8513}&\textcolor{red}{29.41}&\textcolor{red}{0.8132}&\textcolor{red}{29.57}&\textcolor{red}{0.8770}&\textcolor{red}{35.08}&\textcolor{red}{0.9528}\\
\hdashline
IPT \cite{DBLP:conf/cvpr/Chen000DLMX0021}&$\times3$&ImageNet&34.81&-&30.85&-&29.38&-&29.49&-&-&-\\
SwinIR \cite{DBLP:conf/iccvw/LiangCSZGT21}&$\times3$&DIV2K+Flickr2K&\textcolor{blue}{34.97}&0.9318&30.93&\textcolor{red}{0.8534}&29.46&0.8145&29.75&\textcolor{red}{0.8826}&35.12&\textcolor{blue}{0.9537}\\
MaxSR (Ours) &$\times3$&DIV2K+Flickr2K&\textcolor{blue}{34.97}&\textcolor{blue}{0.9320}&\textcolor{blue}{30.97}&\textcolor{blue}{0.8528}&\textcolor{blue}{29.48}&\textcolor{blue}{0.8151}&\textcolor{blue}{29.78}&0.8804&\textcolor{blue}{35.22}&0.9535\\
MaxSR+ (Ours) &$\times3$&DIV2K+Flickr2K&\textcolor{red}{35.05}&\textcolor{red}{0.9324}&\textcolor{red}{31.02}&\textcolor{red}{0.8534}&\textcolor{red}{29.51}&\textcolor{red}{0.8155}&\textcolor{red}{29.91}&\textcolor{blue}{0.8819}&\textcolor{red}{35.33}&\textcolor{red}{0.9539}\\
\hline
\hline
EDSR \cite{lim2017enhanced}&$\times4$&DIV2K&32.46&0.8968&28.80&0.7876&27.71&0.7420&26.64&0.8033&31.02&0.9148\\
RDN \cite{zhang2018residual}&$\times4$&DIV2K&32.47&0.8990&28.81&0.7871&27.72&0.7419&26.61&0.8028&31.00&0.9151\\
RCAN \cite{zhang2018image}&$\times4$&DIV2K&32.63&0.9002&28.87&0.7889&27.77&0.7436&26.82&0.8087&31.22&0.9173\\
NLRN \cite{DBLP:conf/nips/LiuWFLH18}&$\times4$&291&31.92&0.8916&28.36&0.7745&27.48&0.7306&25.79&0.7729&-&-\\
RNAN \cite{DBLP:conf/iclr/ZhangLLZF19}&$\times4$&DIV2K&32.49&0.8982&28.83&0.7878&27.72&0.7421&26.61&0.8023&31.09&0.9149\\
SAN \cite{DBLP:conf/cvpr/DaiCZXZ19}&$\times4$&DIV2K&32.64&0.9003&28.92&0.7888&27.78&0.7436&26.79&0.8068&31.18&0.9169\\
OISR-RK3 \cite{DBLP:conf/cvpr/HeMWLY019}&$\times4$&DIV2K&32.53&0.8992&28.86&0.7878&27.75&0.7428&26.79&0.8068&-&-\\
IGNN \cite{DBLP:conf/nips/ZhouZZL20}&$\times4$&DIV2K&32.57&0.8998&28.85&0.7891&27.77&0.7434&26.84&0.8090&31.28&0.9182\\
HAN \cite{niu2020single}&$\times4$&DIV2K&32.64&0.9002&28.90&0.7890&27.80&0.7442&26.85&0.8094&31.42&0.9177\\
NLSN \cite{mei2021image}&$\times4$&DIV2K&32.59&0.9000&28.87&0.7891&27.78&0.7444&26.96&0.8109&31.27&0.9184\\
SwinIR \cite{DBLP:conf/iccvw/LiangCSZGT21}&$\times4$&DIV2K&32.72&0.9021&28.94&0.7914&\textcolor{blue}{27.83}&0.7459&27.07&0.8164&31.67&0.9226\\
ENLCN \cite{DBLP:conf/aaai/XiaHTYLZ22}&$\times4$&DIV2K&32.67&0.9004&28.94&0.7892&27.82&0.7452&27.12&0.8141&31.33&0.9188\\
ELAN \cite{DBLP:conf/eccv/ZhangZGZ22}&$\times4$&DIV2K&\textcolor{blue}{32.75}&\textcolor{blue}{0.9022}&28.96&0.7914&\textcolor{blue}{27.83}&0.7459&27.13&\textcolor{blue}{0.8167}&31.68&0.9226\\
MaxSR (Ours) &$\times4$&DIV2K&32.73&0.9020&\textcolor{blue}{29.00}&\textcolor{blue}{0.7918}&\textcolor{blue}{27.83}&\textcolor{blue}{0.7462}&\textcolor{blue}{27.18}&0.8164&\textcolor{blue}{31.80}&\textcolor{blue}{0.9228}\\
MaxSR+ (Ours) &$\times4$&DIV2K&\textcolor{red}{32.83}&\textcolor{red}{0.9029}&\textcolor{red}{29.09}&\textcolor{red}{0.7934}&\textcolor{red}{27.89}&\textcolor{red}{0.7472}&\textcolor{red}{27.33}&\textcolor{red}{0.8192}&\textcolor{red}{32.01}&\textcolor{red}{0.9244}\\
\hdashline
D-DBPN \cite{haris2018deep}&$\times4$&DIV2K+Flickr2K&32.47&0.8980&28.82&0.7860&27.72&0.7400&26.38&0.7946&30.91&0.9137\\
IPT \cite{DBLP:conf/cvpr/Chen000DLMX0021}&$\times4$&ImageNet&32.64&-&29.01&-&27.82&-&27.26&-&-&-\\
RRDB \cite{DBLP:conf/eccv/WangYWGLDQL18}&$\times4$&DIV2K+Flickr2K&32.73&0.9011&28.99&0.7917&27.85&0.7455&27.03&0.8153&31.66&0.9196\\
SwinIR \cite{DBLP:conf/iccvw/LiangCSZGT21}&$\times4$&DIV2K+Flickr2K&\textcolor{blue}{32.92}&\textcolor{red}{0.9044}&29.09&\textcolor{blue}{0.7950}&\textcolor{blue}{27.92}&\textcolor{blue}{0.7489}&27.45&\textcolor{blue}{0.8254}&32.03&\textcolor{blue}{0.9260}\\
MaxSR (Ours) &$\times4$&DIV2K+Flickr2K&32.88&0.9034&\textcolor{blue}{29.13}&0.7948&\textcolor{blue}{27.92}&\textcolor{blue}{0.7489}&\textcolor{blue}{27.51}&0.8247&\textcolor{blue}{32.16}&0.9259\\
MaxSR+ (Ours) &$\times4$&DIV2K+Flickr2K&\textcolor{red}{32.97}&\textcolor{blue}{0.9042}&\textcolor{red}{29.21}&\textcolor{red}{0.7962}&\textcolor{red}{27.97}&\textcolor{red}{0.7499}&\textcolor{red}{27.67}&\textcolor{red}{0.8273}&\textcolor{red}{32.34}&\textcolor{red}{0.9271}\\
\hline
\hline
EDSR \cite{lim2017enhanced}&$\times8$&DIV2K&26.96&0.7762&24.91&0.6420&24.81&0.5985&22.51&0.6221&24.69&0.7841\\
RCAN \cite{zhang2018image}&$\times8$&DIV2K&27.31&0.7878&25.23&0.6511&24.98&0.6058&23.00&0.6452&25.24&\textcolor{blue}{0.8029}\\
SAN \cite{DBLP:conf/cvpr/DaiCZXZ19}&$\times8$&DIV2K&27.22&0.7829&25.14&0.6476&24.88&0.6011&22.70&0.6314&24.85&0.7906\\
HAN \cite{niu2020single}&$\times8$&DIV2K&27.33&\textcolor{blue}{0.7884}&25.24&0.6510&24.98&0.6059&22.98&0.6437&25.20&0.8011\\
SwinIR \cite{DBLP:conf/iccvw/LiangCSZGT21}&$\times8$&DIV2K&\textcolor{blue}{27.37}&0.7877&25.26&\textcolor{blue}{0.6523}&\textcolor{blue}{24.99}&\textcolor{blue}{0.6063}&23.03&0.6457&25.26&0.8005\\
MaxSR (Ours) &$\times8$&DIV2K&\textcolor{blue}{27.37}&0.7867&\textcolor{blue}{25.30}&0.6515&24.96&0.6049&\textcolor{blue}{23.11}&\textcolor{blue}{0.6479}&\textcolor{blue}{25.33}&0.8020\\
MaxSR+ (Ours) &$\times8$&DIV2K&\textcolor{red}{27.42}&\textcolor{red}{0.7895}&\textcolor{red}{25.37}&\textcolor{red}{0.6537}&\textcolor{red}{25.01}&\textcolor{red}{0.6064}&\textcolor{red}{23.20}&\textcolor{red}{0.6515}&\textcolor{red}{25.48}&\textcolor{red}{0.8055}\\
\hdashline
D-DBPN \cite{haris2018deep}&$\times8$&DIV2K+Flickr2K&27.21&0.7840&25.13&0.6480&24.88&0.6010&22.73&0.6312&25.14&0.7987\\
SwinIR \cite{DBLP:conf/iccvw/LiangCSZGT21}&$\times8$&DIV2K+Flickr2K&\textcolor{red}{27.55}&\textcolor{red}{0.7941}&\textcolor{blue}{25.46}&\textcolor{blue}{0.6568}&25.04&0.6092&23.17&0.6547&25.55&0.8132\\
MaxSR (Ours) &$\times8$&DIV2K+Flickr2K&27.38&0.7894&25.39&0.6556&\textcolor{blue}{25.06}&\textcolor{blue}{0.6099}&\textcolor{blue}{23.36}&\textcolor{blue}{0.6607}&\textcolor{blue}{25.69}&\textcolor{blue}{0.8142}\\
MaxSR+ (Ours) &$\times8$&DIV2K+Flickr2K&\textcolor{blue}{27.53}&\textcolor{blue}{0.7938}&\textcolor{red}{25.48}&\textcolor{red}{0.6583}&\textcolor{red}{25.11}&\textcolor{red}{0.6114}&\textcolor{red}{23.47}&\textcolor{red}{0.6642}&\textcolor{red}{25.86}&\textcolor{red}{0.8174}\\
\hline
\end{tabularx}
\\
}
\end{center}
}
\end{table*}
\begin{table*}[ht]
{
\caption{Average PSNR/SSIM Metrics for Scale Factors \(\times2\), \(\times3\), \(\times4\) of Compared Methods for Lightwight Image SR.
Best is in \textcolor{red}{Red}, and Second Best is in \textcolor{blue}{Blue}.
}
\label{tab:lightweight_pnsr_ssim_table}
\scriptsize
\begin{center}
{
\begin{tabularx}{.96\linewidth}{|X|c|c|c|c|c|c|c|c|c|c|c|c|}
\hline
\multirow{2}*{Method}&\multirow{2}*{Scale}&\multirow{2}*{\#Params}&\multicolumn{2}{|c|}{Set5}&\multicolumn{2}{|c|}{Set14}&\multicolumn{2}{|c|}{B100}&\multicolumn{2}{|c|}{Urban100}&\multicolumn{2}{|c|}{Manga109}\\
\cline{4-13}
&&&PSNR&SSIM&PSNR&SSIM&PSNR&SSIM&PSNR&SSIM&PSNR&SSIM\\
\hline
\hline
DRRN \cite{DBLP:conf/cvpr/TaiY017}&$\times2$&0.30M&37.74&0.9591&33.23&0.9136&32.05&0.8973&31.23&0.9188&37.92&0.9760\\
CARN \cite{DBLP:conf/eccv/AhnKS18}&$\times2$&1.59M&37.76&0.9590&33.52&0.9166&32.09&0.8978&31.92&0.9256&38.36&0.9765\\
IDN \cite{hui2018fast}&$\times2$&0.55M&37.83&0.9600&33.30&0.9148&32.08&0.8985&31.27&0.9196&38.88&0.9774\\
IMDN \cite{2019Lightweight}&$\times2$&0.69M&38.00&0.9605&33.63&0.9177&32.19&0.8999&32.17&0.9283&38.01&0.9749\\
FALSR-A \cite{DBLP:conf/icpr/Chu0MXL20}&$\times2$&1.02M&37.82&0.9595&33.55&0.9168&32.12&0.8987&31.93&0.9256&-&-\\
FALSR-C \cite{DBLP:conf/icpr/Chu0MXL20}&$\times2$&0.41M&37.66&0.9586&33.26&0.9140&31.96&0.8965&31.24&0.9187&-&-\\
AAF-SD \cite{DBLP:conf/accv/WangWZYFC20}&$\times2$&0.31M&37.91&0.9602&33.45&0.9164&32.08&0.8986&31.79&0.9246&38.52&0.9767\\
MAFFSRN \cite{DBLP:conf/eccv/MuqeetHYKKB20}&$\times2$&0.40M&37.97&0.9603&33.49&0.9170&32.14&0.8994&31.96&0.9268&-&-\\
PAN \cite{DBLP:conf/eccv/ZhaoKHQD20}&$\times2$&0.26M&38.00&0.9605&33.59&0.9181&32.18&0.8997&32.01&0.9273&38.70&0.9773\\
LAPAR-A \cite{DBLP:conf/nips/LiZQJLJ20}&$\times2$&0.55M&38.01&0.9605&33.62&0.9183&32.19&0.8999&32.10&0.9283&38.67&0.9772\\
RFDN \cite{DBLP:conf/eccv/LiuTW20}&$\times2$&0.53M&38.05&0.9606&33.68&0.9184&32.16&0.8994&32.12&0.9278&38.88&0.9773\\
AAF-L \cite{DBLP:conf/accv/WangWZYFC20}&$\times2$&1.36M&38.09&0.9607&33.78&0.9192&32.23&0.9002&32.46&0.9313&38.95&0.9772\\
A-CubeNet \cite{DBLP:conf/mm/HangLYC020}&$\times2$&1.38M&38.12&0.9609&33.73&0.9191&32.26&0.9007&32.39&0.9308&38.88&0.9776\\
LatticeNet \cite{DBLP:conf/eccv/LuoXZQLF20}&$\times2$&0.76M&\textcolor{blue}{38.15}&0.9610&33.78&0.9193&32.25&0.9005&32.43&0.9302&-&-\\
SwinIR-light \cite{DBLP:conf/iccvw/LiangCSZGT21}&$\times2$&0.88M&38.14&\textcolor{blue}{0.9611}&33.86&0.9206&\textcolor{red}{32.31}&\textcolor{red}{0.9012}&\textcolor{blue}{32.76}&\textcolor{blue}{0.9340}&\textcolor{blue}{39.12}&\textcolor{blue}{0.9783}\\
ELAN-light \cite{DBLP:conf/eccv/ZhangZGZ22}&$\times2$&0.58M&\textcolor{red}{38.17}&\textcolor{blue}{0.9611}&\textcolor{blue}{33.94}&\textcolor{blue}{0.9207}&\textcolor{blue}{32.30}&\textcolor{red}{0.9012}&\textcolor{blue}{32.76}&\textcolor{blue}{0.9340}&39.11&0.9782\\
MaxSR-light (Ours) &$\times2$&0.90M&38.12&\textcolor{red}{0.9612}&\textcolor{red}{34.13}&\textcolor{red}{0.9223}&\textcolor{blue}{32.30}&\textcolor{blue}{0.9011}&\textcolor{red}{32.88}&\textcolor{red}{0.9341}&\textcolor{red}{39.35}&\textcolor{red}{0.9785}\\
\hline
\hline
DRRN \cite{DBLP:conf/cvpr/TaiY017}&$\times3$&0.30M&34.03&0.9244&29.96&0.8349&28.95&0.8004&27.53&0.8378&32.74&0.9390\\
CARN \cite{DBLP:conf/eccv/AhnKS18}&$\times3$&1.59M&34.29&0.9255&30.29&0.8407&29.06&0.8034&28.06&0.8493&33.50&0.9440\\
IDN \cite{hui2018fast}&$\times3$&0.55M&34.11&0.9253&29.99&0.8354&28.95&0.8013&27.42&0.8359&32.71&0.9381\\
IMDN \cite{2019Lightweight}&$\times3$&0.70M&34.36&0.9270&30.32&0.8417&29.09&0.8046&28.17&0.8519&33.61&0.9445\\
AAF-SD \cite{DBLP:conf/accv/WangWZYFC20}&$\times3$&0.32M&34.23&0.9259&30.22&0.8395&29.01&0.8028&27.91&0.8465&33.29&0.9424\\
MAFFSRN \cite{DBLP:conf/eccv/MuqeetHYKKB20}&$\times3$&0.42M&34.32&0.9269&30.35&0.8429&29.09&0.8052&28.13&0.8521&-&-\\
LAPAR-A \cite{DBLP:conf/nips/LiZQJLJ20}&$\times3$&0.59M&34.36&0.9267&30.34&0.8421&29.11&0.8054&28.15&0.8523&33.51&0.9441\\
PAN \cite{DBLP:conf/eccv/ZhaoKHQD20}&$\times3$&0.26M&34.40&0.9272&30.36&0.8422&29.11&0.8049&28.11&0.8511&33.61&0.9448\\
RFDN \cite{DBLP:conf/eccv/LiuTW20}&$\times3$&0.54M&34.41&0.9273&30.34&0.8420&29.09&0.8050&28.21&0.8525&33.67&0.9449\\
LatticeNet \cite{DBLP:conf/eccv/LuoXZQLF20}&$\times3$&0.77M&34.53&0.9281&30.39&0.8424&29.15&0.8059&28.33&0.8538&-&-\\
A-CubeNet \cite{DBLP:conf/mm/HangLYC020}&$\times3$&1.56M&34.53&0.9281&30.45&0.8441&29.17&0.8068&28.38&0.8568&33.90&0.9466\\
AAF-L \cite{DBLP:conf/accv/WangWZYFC20}&$\times3$&1.37M&34.54&0.9283&30.41&0.8436&29.14&0.8062&28.40&0.8574&33.83&0.9463\\
SwinIR-light \cite{DBLP:conf/iccvw/LiangCSZGT21}&$\times3$&0.89M&\textcolor{red}{34.62}&\textcolor{red}{0.9289}&\textcolor{blue}{30.54}&\textcolor{red}{0.8463}&29.20&\textcolor{blue}{0.8082}&28.66&\textcolor{blue}{0.8624}&33.98&\textcolor{blue}{0.9478}\\
ELAN-light \cite{DBLP:conf/eccv/ZhangZGZ22}&$\times3$&0.59M&\textcolor{blue}{34.61}&\textcolor{blue}{0.9288}&\textcolor{red}{30.55}&\textcolor{red}{0.8463}&\textcolor{blue}{29.21}&0.8081&\textcolor{blue}{28.69}&\textcolor{blue}{0.8624}&\textcolor{blue}{34.00}&\textcolor{blue}{0.9478}\\
MaxSR-light (Ours) &$\times3$&1.01M&34.59&\textcolor{red}{0.9289}&\textcolor{blue}{30.54}&\textcolor{blue}{0.8457}&\textcolor{red}{29.24}&\textcolor{red}{0.8090}&\textcolor{red}{28.75}&\textcolor{red}{0.8627}&\textcolor{red}{34.28}&\textcolor{red}{0.9485}\\
\hline
\hline
DRRN \cite{DBLP:conf/cvpr/TaiY017}&$\times4$&0.30M&31.68&0.8888&28.21&0.7720&27.38&0.7284&25.44&0.7638&29.46&0.8960\\
CARN \cite{DBLP:conf/eccv/AhnKS18}&$\times4$&1.59M&32.13&0.8937&28.60&0.7806&27.58&0.7349&26.07&0.7837&30.47&0.9084\\
IDN \cite{hui2018fast}&$\times4$&0.55M&31.82&0.8903&28.25&0.7730&27.41&0.7297&25.41&0.7632&29.41&0.8942\\
IMDN \cite{2019Lightweight}&$\times4$&0.72M&32.21&0.8948&28.58&0.7811&27.56&0.7353&26.04&0.7838&30.45&0.9075\\
AAF-SD \cite{DBLP:conf/accv/WangWZYFC20}&$\times4$&0.32M&32.06&0.8928&28.47&0.7790&27.48&0.7373&25.80&0.7767&30.16&0.9038\\
PAN \cite{DBLP:conf/eccv/ZhaoKHQD20}&$\times4$&0.27M&32.13&0.8948&28.61&0.7822&27.59&0.7363&26.11&0.7854&30.51&0.9095\\
LAPAR-A \cite{DBLP:conf/nips/LiZQJLJ20}&$\times4$&0.66M&32.15&0.8944&28.61&0.7818&27.61&0.7366&26.14&0.7871&30.42&0.9074\\
MAFFSRN \cite{DBLP:conf/eccv/MuqeetHYKKB20}&$\times4$&0.44M&32.18&0.8948&28.58&0.7812&27.57&0.7361&26.04&0.7848&-&-\\
RFDN \cite{DBLP:conf/eccv/LiuTW20}&$\times4$&0.55M&32.24&0.8952&28.61&0.7819&27.57&0.7360&26.11&0.7858&30.58&0.9089\\
LatticeNet \cite{DBLP:conf/eccv/LuoXZQLF20}&$\times4$&0.78M&32.30&0.8962&28.68&0.7830&27.62&0.7367&26.25&0.7873&-&-\\
AAF-L \cite{DBLP:conf/accv/WangWZYFC20}&$\times4$&1.37M&32.32&0.8964&28.67&0.7839&27.62&0.7379&26.32&0.7931&30.72&0.9115\\
A-CubeNet \cite{DBLP:conf/mm/HangLYC020}&$\times4$&1.52M&32.32&0.8969&28.72&0.7847&27.65&0.7382&26.27&0.7913&30.81&0.9114\\
SwinIR-light \cite{DBLP:conf/iccvw/LiangCSZGT21}&$\times4$&0.90M&\textcolor{red}{32.44}&\textcolor{blue}{0.8976}&\textcolor{blue}{28.77}&\textcolor{red}{0.7858}&\textcolor{red}{27.69}&\textcolor{red}{0.7406}&26.47&0.7980&\textcolor{blue}{30.92}&\textcolor{blue}{0.9151}\\
ELAN-light \cite{DBLP:conf/eccv/ZhangZGZ22}&$\times4$&0.60M&\textcolor{blue}{32.43}&0.8975&\textcolor{red}{28.78}&\textcolor{red}{0.7858}&\textcolor{red}{27.69}&\textcolor{red}{0.7406}&\textcolor{blue}{26.54}&\textcolor{blue}{0.7982}&\textcolor{blue}{30.92}&0.9150\\
MaxSR-light (Ours) &$\times4$&0.99M&32.41&\textcolor{red}{0.8980}&\textcolor{blue}{28.77}&\textcolor{blue}{0.7854}&\textcolor{blue}{27.67}&\textcolor{blue}{0.7395}&\textcolor{red}{26.58}&\textcolor{red}{0.8002}&\textcolor{red}{31.12}&\textcolor{red}{0.9154}\\
\hline
\end{tabularx}
\\
}
\end{center}
}
\end{table*}
\begin{figure*}[ht]
\centering
\scriptsize
\begin{minipage}{0.32\linewidth}
\centering
\begin{subfigure}[t]{\linewidth}
\centering\includegraphics[width=\linewidth, height=124pt]{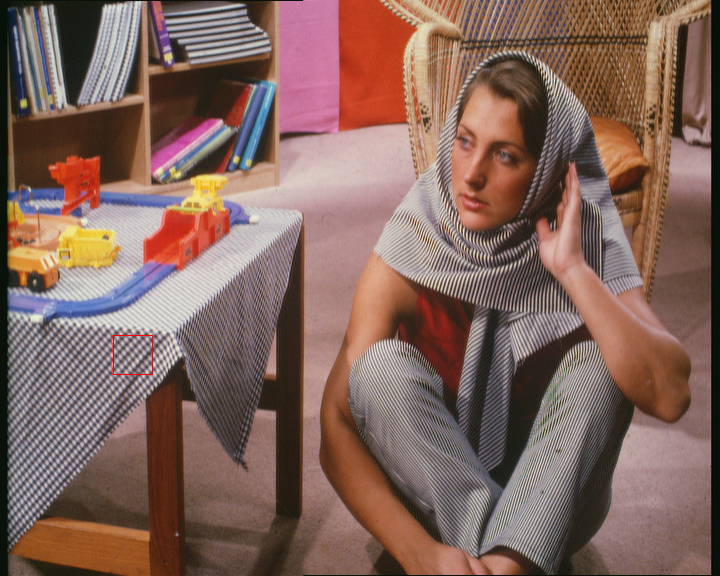}
\caption*{Set14 ($\times4$)\\barbara}
\end{subfigure}
\end{minipage}
\centering
\begin{minipage}{0.64\linewidth}
\centering
\begin{subfigure}[t]{.192\linewidth}
\centering\includegraphics[width=\linewidth, height=30pt]{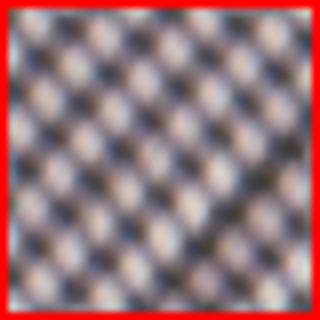}
\caption*{HR\\PSNR/SSIM}
\end{subfigure}
\begin{subfigure}[t]{.192\linewidth}
\centering\includegraphics[width=\linewidth, height=30pt]{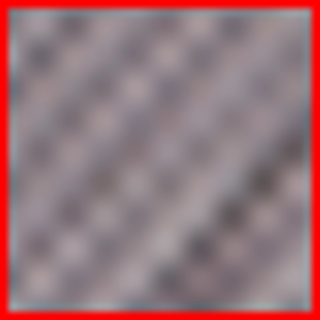}
\caption*{Bicubic\\25.15/0.6863}
\end{subfigure}
\begin{subfigure}[t]{.192\linewidth}
\centering\includegraphics[width=\linewidth, height=30pt]{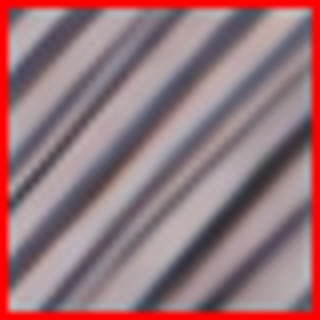}
\caption*{EDSR\\26.42/0.7645}
\end{subfigure}
\begin{subfigure}[t]{.192\linewidth}
\centering\includegraphics[width=\linewidth, height=30pt]{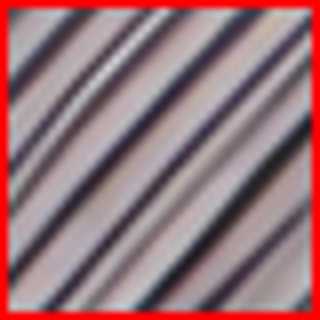}
\caption*{RDN\\26.07/0.7598}
\end{subfigure}
\begin{subfigure}[t]{.192\linewidth}
\centering\includegraphics[width=\linewidth, height=30pt]{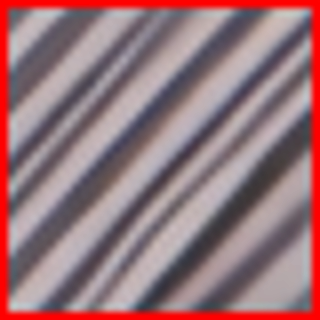}
\caption*{D-DBPN\\26.22/0.7585}
\end{subfigure}

\begin{subfigure}[t]{.192\linewidth}
\centering\includegraphics[width=\linewidth, height=30pt]{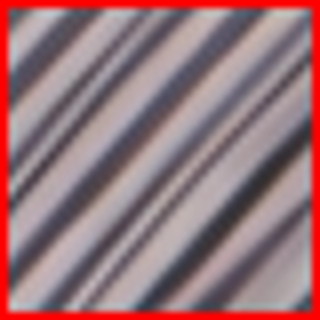}
\caption*{RCAN\\26.25/0.7620}
\end{subfigure}
\begin{subfigure}[t]{.192\linewidth}
\centering\includegraphics[width=\linewidth, height=30pt]{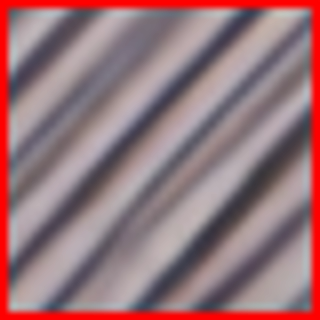}
\caption*{RNAN\\26.47/0.7644}
\end{subfigure}
\begin{subfigure}[t]{.192\linewidth}
\centering\includegraphics[width=\linewidth, height=30pt]{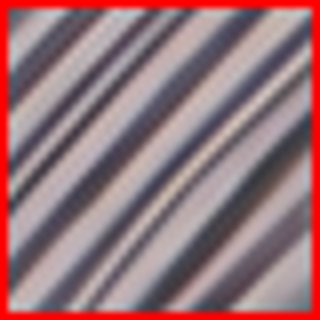}
\caption*{SAN\\26.25/0.7614}
\end{subfigure}
\begin{subfigure}[t]{.192\linewidth}
\centering\includegraphics[width=\linewidth, height=30pt]{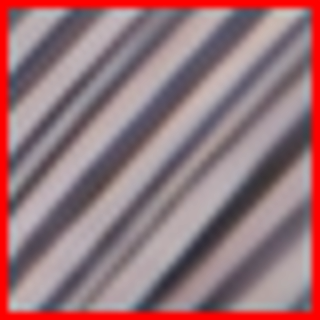}
\caption*{OISR-RK3\\26.41/0.7647}
\end{subfigure}
\begin{subfigure}[t]{.192\linewidth}
\centering\includegraphics[width=\linewidth, height=30pt]{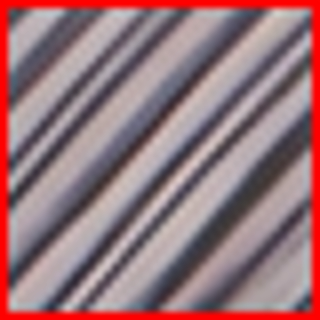}
\caption*{IGNN\\26.27/0.7644}
\end{subfigure}

\begin{subfigure}[t]{.192\linewidth}
\centering\includegraphics[width=\linewidth, height=30pt]{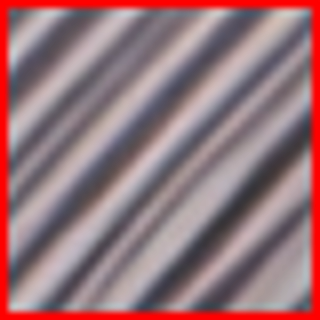}
\caption*{HAN\\26.49/0.7678}
\end{subfigure}
\begin{subfigure}[t]{.192\linewidth}
\centering\includegraphics[width=\linewidth, height=30pt]{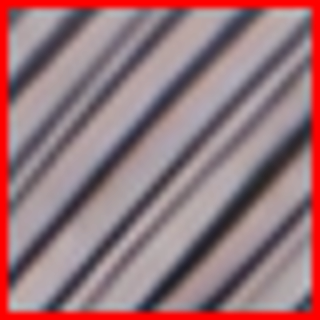}
\caption*{NLSN\\26.27/0.7647}
\end{subfigure}
\begin{subfigure}[t]{.192\linewidth}
\centering\includegraphics[width=\linewidth, height=30pt]{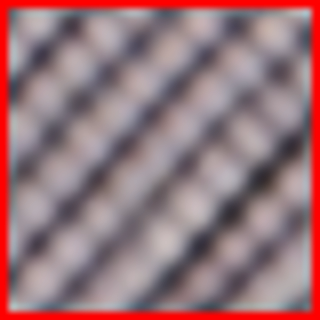}
\caption*{IPT\\\textcolor{blue}{26.67}/\textcolor{blue}{0.7741}}
\end{subfigure}
\begin{subfigure}[t]{.192\linewidth}
\centering\includegraphics[width=\linewidth, height=30pt]{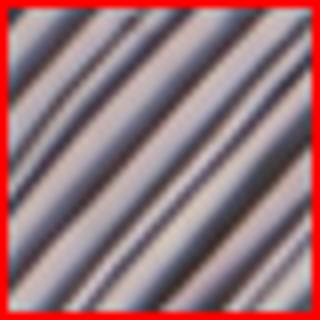}
\caption*{SwinIR\\26.31/0.7707}
\end{subfigure}
\begin{subfigure}[t]{.192\linewidth}
\centering\includegraphics[width=\linewidth, height=30pt]{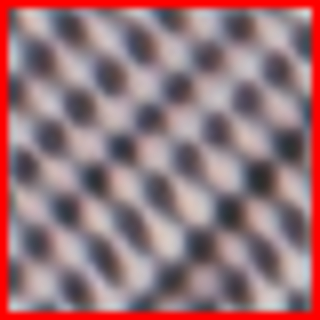}
\caption*{MaxSR\\\textcolor{red}{26.89}/\textcolor{red}{0.7818}}
\end{subfigure}
\end{minipage}
\begin{minipage}{.32\linewidth}
\centering
\begin{subfigure}{\linewidth}
\centering\includegraphics[width=\linewidth, height=124pt]{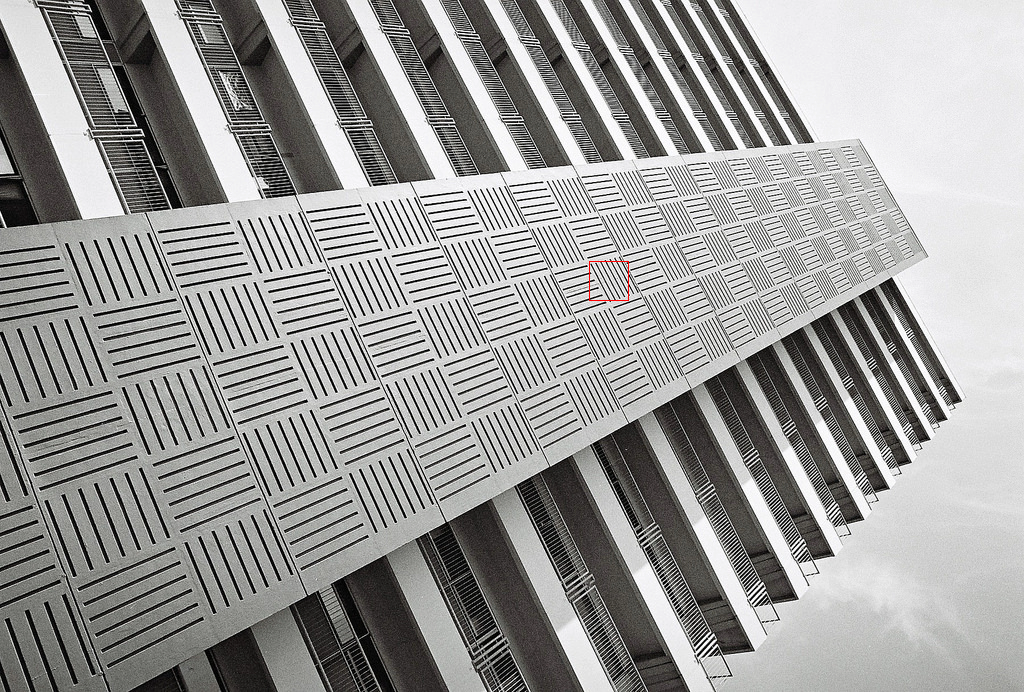}
\caption*{Urban100 ($\times4$)\\img092}
\end{subfigure}
\end{minipage}
\centering
\begin{minipage}{.64\linewidth}
\centering
\begin{subfigure}{.192\linewidth}
\centering\includegraphics[width=\linewidth, height=30pt]{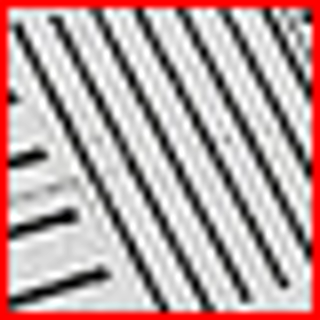}
\caption*{HR\\PSNR/SSIM}
\end{subfigure}
\begin{subfigure}{.192\linewidth}
\centering\includegraphics[width=\linewidth, height=30pt]{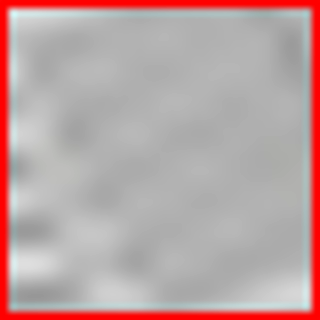}
\caption*{Bicubic\\16.58/0.4372}
\end{subfigure}
\begin{subfigure}{.192\linewidth}
\centering\includegraphics[width=\linewidth, height=30pt]{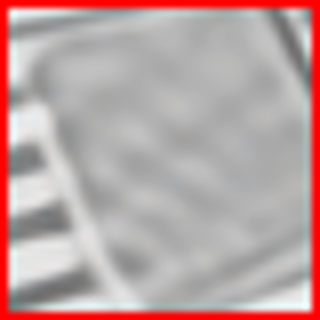}
\caption*{EDSR\\19.15/0.6778}
\end{subfigure}
\begin{subfigure}{.192\linewidth}
\centering\includegraphics[width=\linewidth, height=30pt]{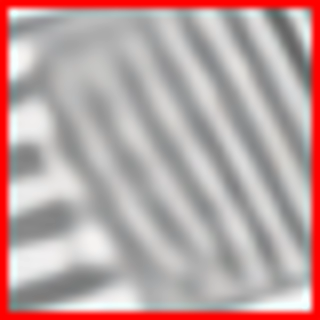}
\caption*{RDN\\19.17/0.6762}
\end{subfigure}
\begin{subfigure}{.192\linewidth}
\centering\includegraphics[width=\linewidth, height=30pt]{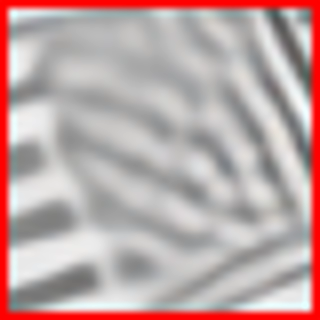}
\caption*{D-DBPN\\18.90/0.6591}
\end{subfigure}

\begin{subfigure}{.192\linewidth}
\centering\includegraphics[width=\linewidth, height=30pt]{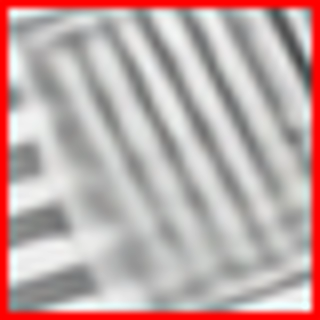}
\caption*{RCAN\\19.66/0.6962}
\end{subfigure}
\begin{subfigure}{.192\linewidth}
\centering\includegraphics[width=\linewidth, height=30pt]{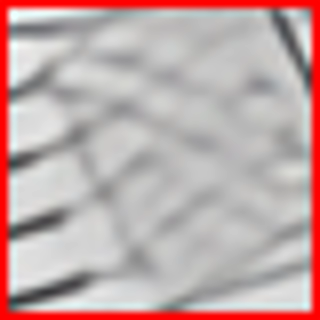}
\caption*{RNAN\\19.51/0.6850}
\end{subfigure}
\begin{subfigure}{.192\linewidth}
\centering\includegraphics[width=\linewidth, height=30pt]{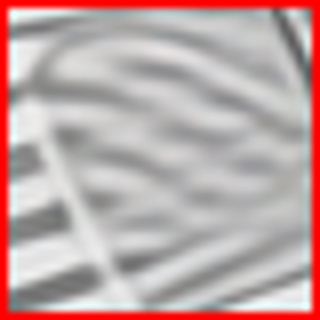}
\caption*{SAN\\19.52/0.6936}
\end{subfigure}
\begin{subfigure}{.192\linewidth}
\centering\includegraphics[width=\linewidth, height=30pt]{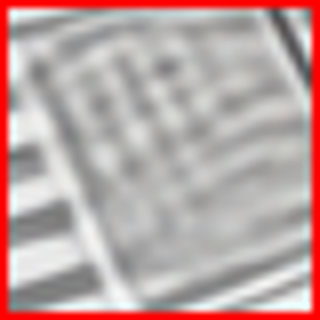}
\caption*{OISR-RK3\\19.18/0.6786}
\end{subfigure}
\begin{subfigure}{.192\linewidth}
\centering\includegraphics[width=\linewidth, height=30pt]{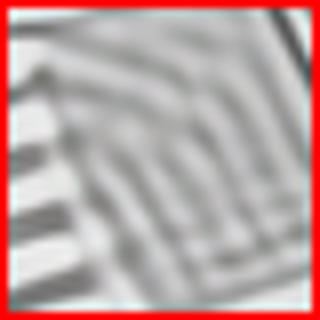}
\caption*{IGNN\\19.78/0.7010}
\end{subfigure}

\begin{subfigure}{.192\linewidth}
\centering\includegraphics[width=\linewidth, height=30pt]{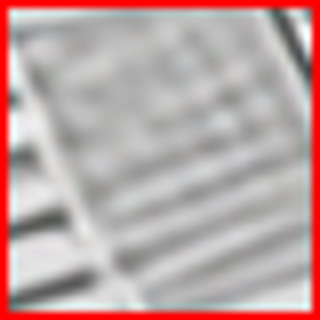}
\caption*{HAN\\19.57/0.6879}
\end{subfigure}
\begin{subfigure}{.192\linewidth}
\centering\includegraphics[width=\linewidth, height=30pt]{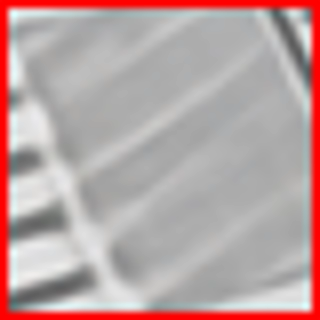}
\caption*{NLSN\\19.54/0.6937}
\end{subfigure}
\begin{subfigure}{.192\linewidth}
\centering\includegraphics[width=\linewidth, height=30pt]{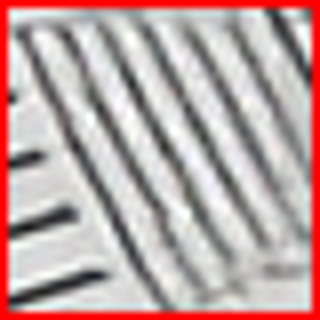}
\caption*{IPT\\\textcolor{blue}{20.38}/\textcolor{blue}{0.7257}}
\end{subfigure}
\begin{subfigure}{.192\linewidth}
\centering\includegraphics[width=\linewidth, height=30pt]{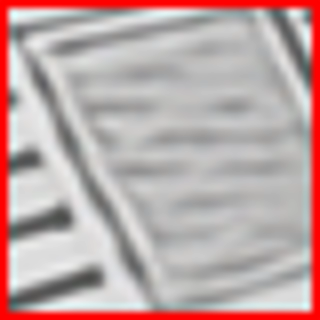}
\caption*{SwinIR\\19.87/0.7079}
\end{subfigure}
\begin{subfigure}{.192\linewidth}
\centering\includegraphics[width=\linewidth, height=30pt]{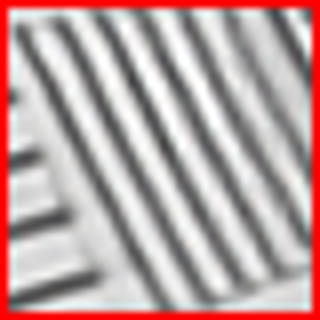}
\caption*{MaxSR\\\textcolor{red}{20.51}/\textcolor{red}{0.7297}}
\end{subfigure}
\end{minipage}
\caption{
Visual Results of Compared Methods for Classical Image SR.
}
\label{fig:visual_comparison_classical}
\end{figure*}
\begin{figure*}[ht]
\centering
\begin{minipage}{.32\linewidth}
\centering
\begin{subfigure}{\linewidth}
\centering\includegraphics[width=\linewidth, height=77pt]{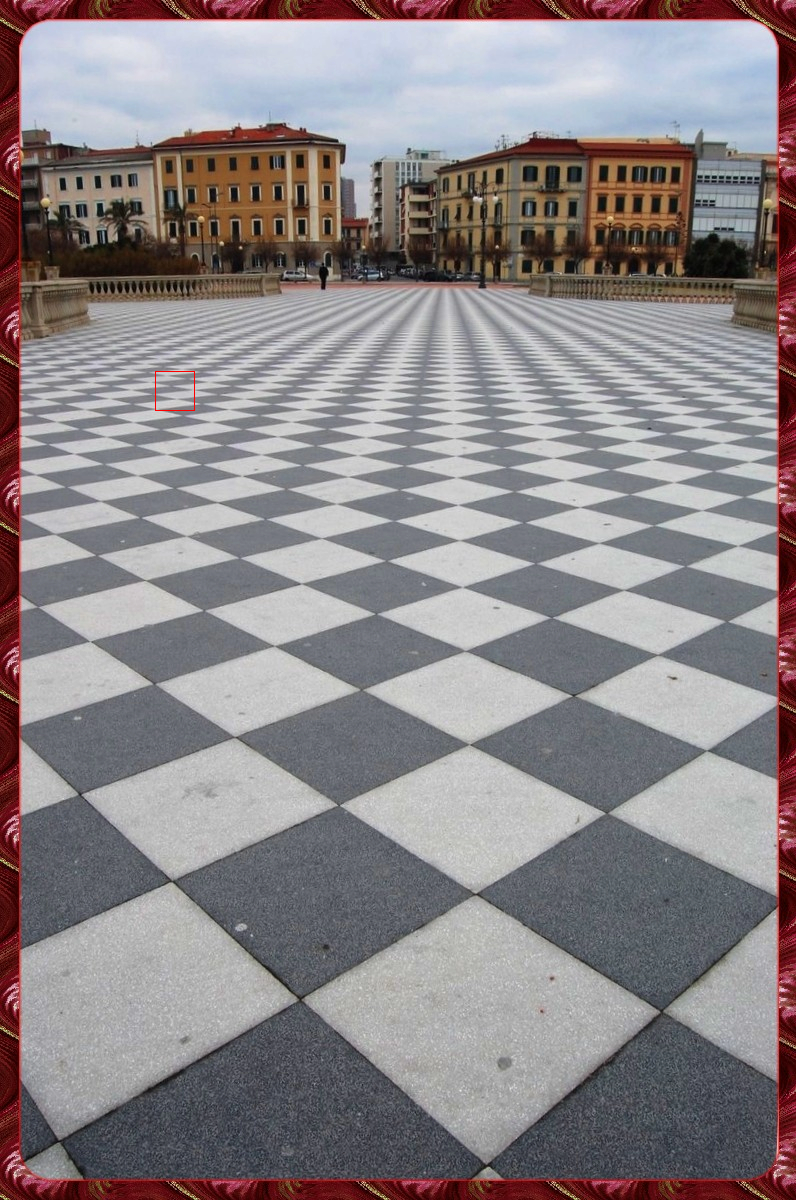}
\caption*{Urban100 ($\times4$)\\img021}
\end{subfigure}
\end{minipage}
\centering
\begin{minipage}{.64\linewidth}
\centering
\begin{subfigure}{.192\linewidth}
\centering\includegraphics[width=\linewidth, height=30pt]{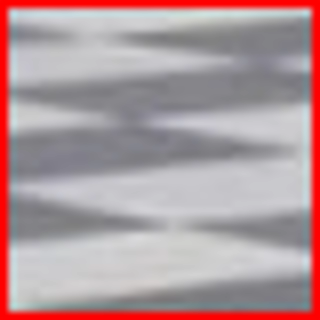}
\caption*{HR\\PSNR/SSIM}
\end{subfigure}
\begin{subfigure}{.192\linewidth}
\centering\includegraphics[width=\linewidth, height=30pt]{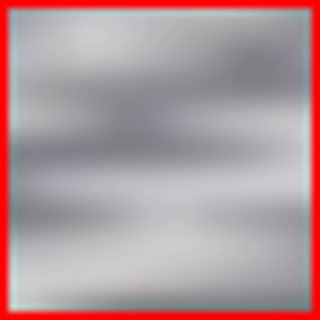}
\caption*{Bicubic\\27.73/0.7198}
\end{subfigure}
\begin{subfigure}{.192\linewidth}
\centering\includegraphics[width=\linewidth, height=30pt]{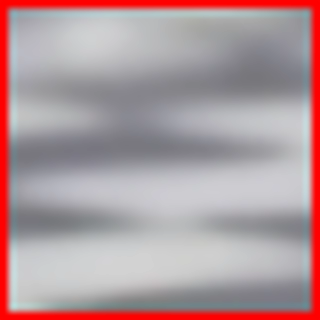}
\caption*{CARN\\29.65/0.7866}
\end{subfigure}
\begin{subfigure}{.192\linewidth}
\centering\includegraphics[width=\linewidth, height=30pt]{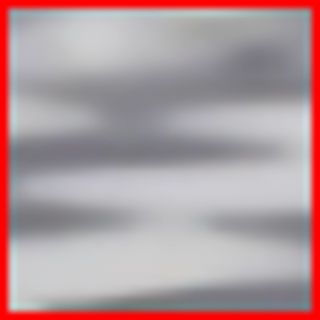}
\caption*{IMDN\\29.64/0.7860}
\end{subfigure}
\begin{subfigure}{.192\linewidth}
\centering\includegraphics[width=\linewidth, height=30pt]{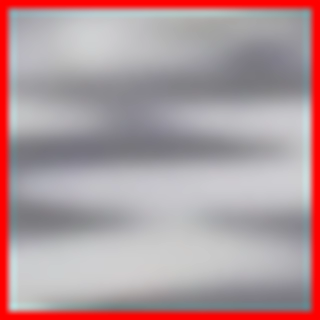}
\caption*{PAN\\29.72/0.7876}
\end{subfigure}

\begin{subfigure}{.192\linewidth}
\centering\includegraphics[width=\linewidth, height=30pt]{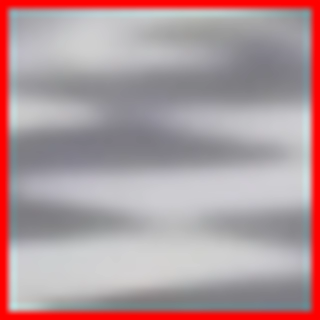}
\caption*{LAPAR-A\\29.80/0.7896}
\end{subfigure}
\begin{subfigure}{.192\linewidth}
\centering\includegraphics[width=\linewidth, height=30pt]{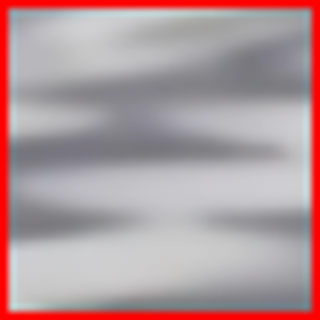}
\caption*{AAF-L\\29.84/0.7915}
\end{subfigure}
\begin{subfigure}{.192\linewidth}
\centering\includegraphics[width=\linewidth, height=30pt]{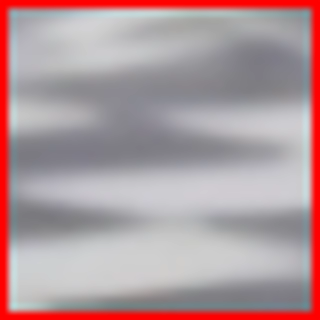}
\caption*{A-CubeNet\\29.75/0.7890}
\end{subfigure}
\begin{subfigure}{.192\linewidth}
\centering\includegraphics[width=\linewidth, height=30pt]{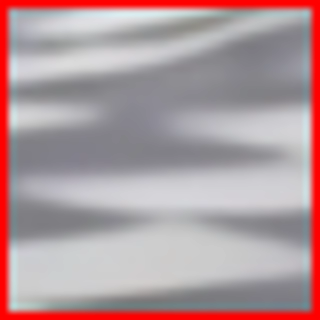}
\caption*{SwinIR-light\\\textcolor{blue}{29.89}/\textcolor{blue}{0.7936}}
\end{subfigure}
\begin{subfigure}{.192\linewidth}
\centering\includegraphics[width=\linewidth, height=30pt]{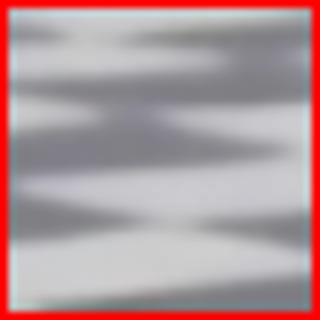}
\caption*{MaxSR-light\\\textcolor{red}{29.97}/\textcolor{red}{0.7959}}
\end{subfigure}
\end{minipage}
\begin{minipage}{.32\linewidth}
\centering
\begin{subfigure}{\linewidth}
\centering\includegraphics[width=\linewidth, height=77pt]{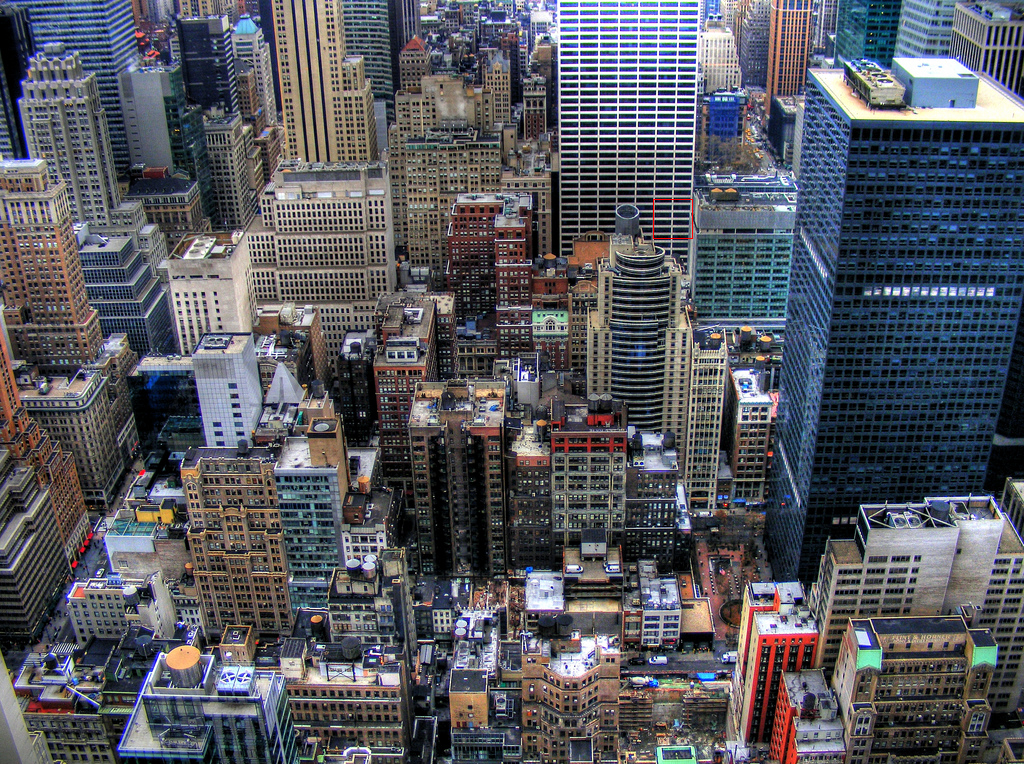}
\caption*{Urban100 ($\times4$)\\img073}
\end{subfigure}
\end{minipage}
\centering
\begin{minipage}{.64\linewidth}
\centering
\begin{subfigure}{.192\linewidth}
\centering\includegraphics[width=\linewidth, height=30pt]{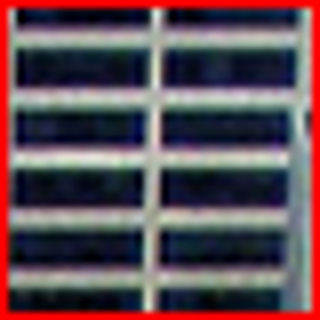}
\caption*{HR\\PSNR/SSIM}
\end{subfigure}
\begin{subfigure}{.192\linewidth}
\centering\includegraphics[width=\linewidth, height=30pt]{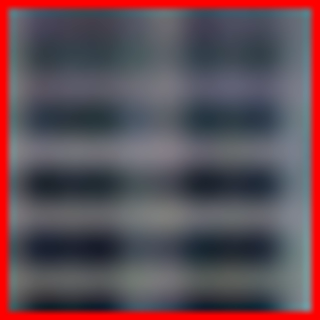}
\caption*{Bicubic\\19.47/0.4374}
\end{subfigure}
\begin{subfigure}{.192\linewidth}
\centering\includegraphics[width=\linewidth, height=30pt]{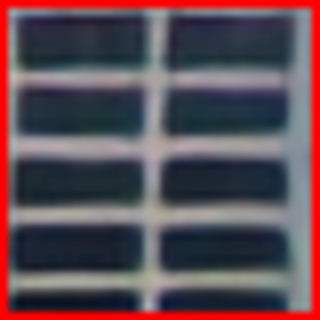}
\caption*{CARN\\20.12/0.5668}
\end{subfigure}
\begin{subfigure}{.192\linewidth}
\centering\includegraphics[width=\linewidth, height=30pt]{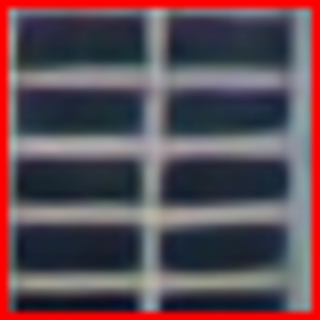}
\caption*{IMDN\\20.09/0.5573}
\end{subfigure}
\begin{subfigure}{.192\linewidth}
\centering\includegraphics[width=\linewidth, height=30pt]{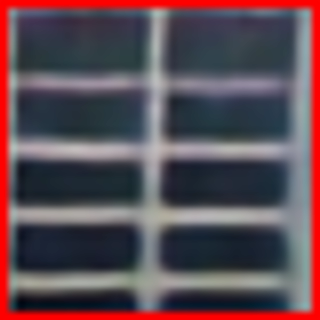}
\caption*{PAN\\20.19/0.5637}
\end{subfigure}

\begin{subfigure}{.192\linewidth}
\centering\includegraphics[width=\linewidth, height=30pt]{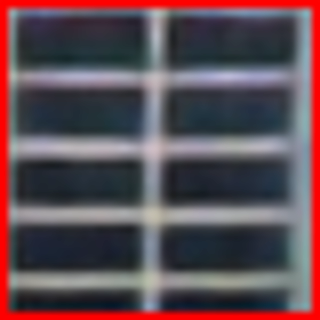}
\caption*{LAPAR-A\\20.09/0.5617}
\end{subfigure}
\begin{subfigure}{.192\linewidth}
\centering\includegraphics[width=\linewidth, height=30pt]{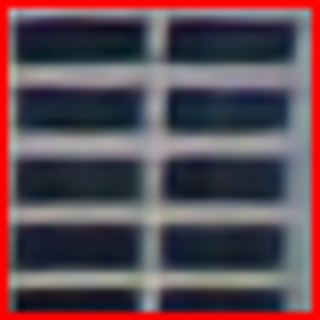}
\caption*{AAF-L\\\textcolor{blue}{20.49}/0.5767}
\end{subfigure}
\begin{subfigure}{.192\linewidth}
\centering\includegraphics[width=\linewidth, height=30pt]{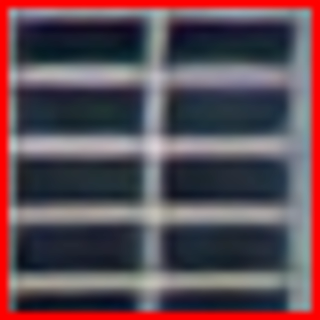}
\caption*{A-CubeNet\\19.61/0.5537}
\end{subfigure}
\begin{subfigure}{.192\linewidth}
\centering\includegraphics[width=\linewidth, height=30pt]{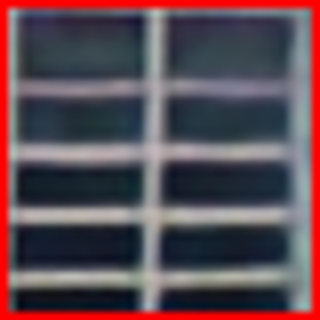}
\caption*{SwinIR-light\\20.09/\textcolor{blue}{0.5908}}
\end{subfigure}
\begin{subfigure}{.192\linewidth}
\centering\includegraphics[width=\linewidth, height=30pt]{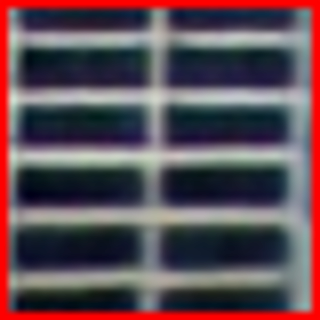}
\caption*{MaxSR-light\\\textcolor{red}{21.16}/\textcolor{red}{0.6068}}
\end{subfigure}
\end{minipage}
\begin{minipage}{.32\linewidth}
\centering
\begin{subfigure}{\linewidth}
\centering\includegraphics[width=\linewidth, height=77pt]{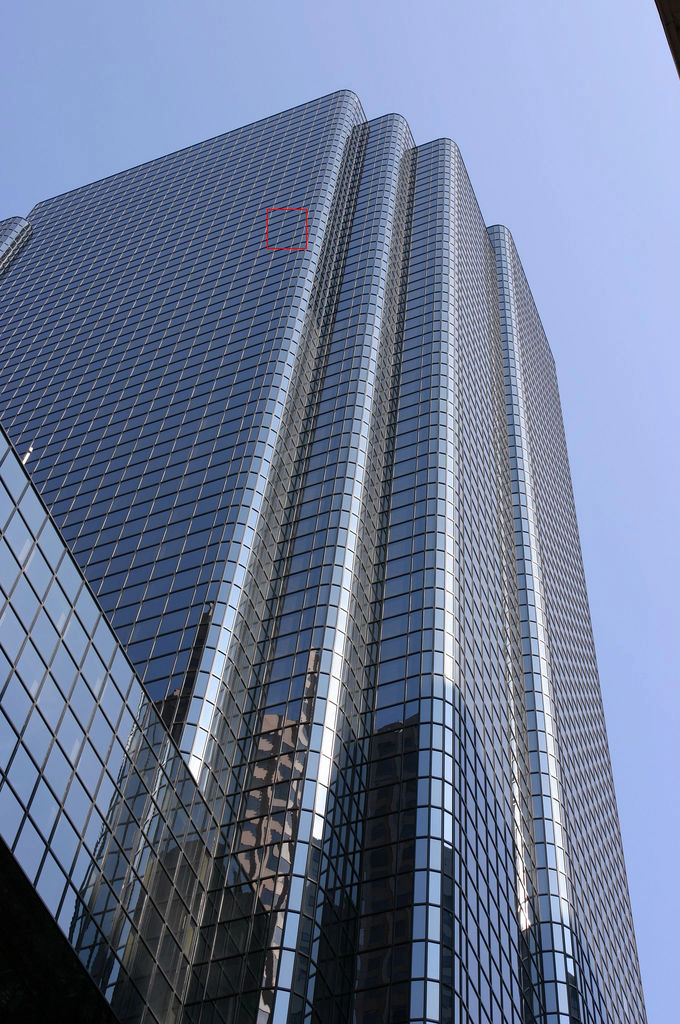}
\caption*{Urban100 ($\times4$)\\img074}
\end{subfigure}
\end{minipage}
\centering
\begin{minipage}{.64\linewidth}
\centering
\begin{subfigure}{.192\linewidth}
\centering\includegraphics[width=\linewidth, height=30pt]{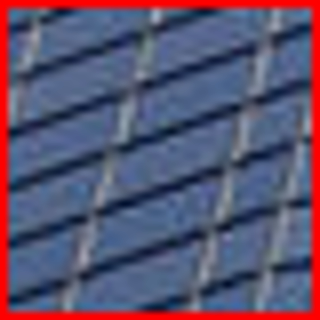}
\caption*{HR\\PSNR/SSIM}
\end{subfigure}
\begin{subfigure}{.192\linewidth}
\centering\includegraphics[width=\linewidth, height=30pt]{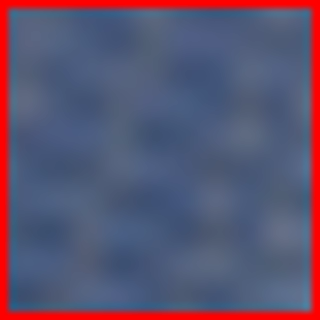}
\caption*{Bicubic\\22.16/0.5551}
\end{subfigure}
\begin{subfigure}{.192\linewidth}
\centering\includegraphics[width=\linewidth, height=30pt]{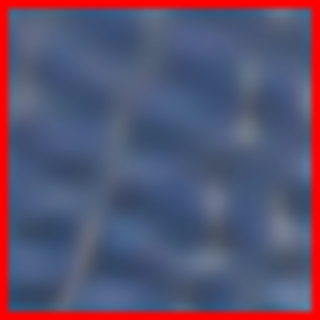}
\caption*{CARN\\23.59/0.6976}
\end{subfigure}
\begin{subfigure}{.192\linewidth}
\centering\includegraphics[width=\linewidth, height=30pt]{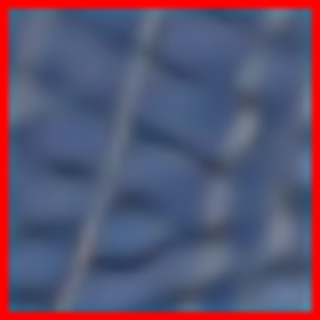}
\caption*{IMDN\\23.75/0.7092}
\end{subfigure}
\begin{subfigure}{.192\linewidth}
\centering\includegraphics[width=\linewidth, height=30pt]{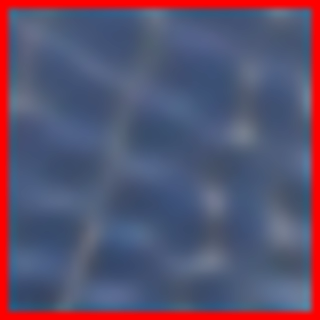}
\caption*{PAN\\23.73/0.7028}
\end{subfigure}

\begin{subfigure}{.192\linewidth}
\centering\includegraphics[width=\linewidth, height=30pt]{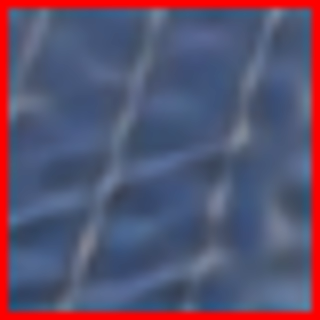}
\caption*{LAPAR-A\\\textcolor{blue}{23.90}/0.7189}
\end{subfigure}
\begin{subfigure}{.192\linewidth}
\centering\includegraphics[width=\linewidth, height=30pt]{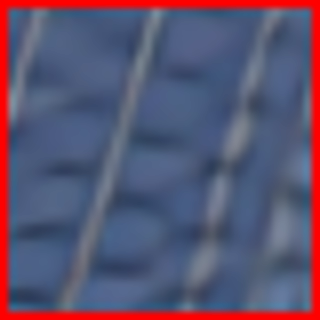}
\caption*{AAF-L\\23.89/\textcolor{blue}{0.7253}}
\end{subfigure}
\begin{subfigure}{.192\linewidth}
\centering\includegraphics[width=\linewidth, height=30pt]{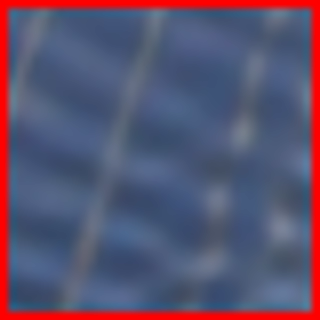}
\caption*{A-CubeNet\\23.76/0.7052}
\end{subfigure}
\begin{subfigure}{.192\linewidth}
\centering\includegraphics[width=\linewidth, height=30pt]{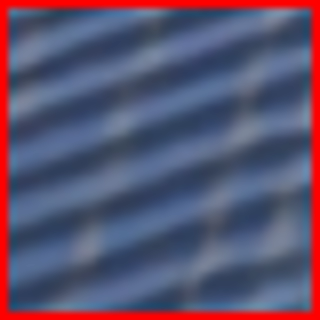}
\caption*{SwinIR-light\\23.78/0.7218}
\end{subfigure}
\begin{subfigure}{.192\linewidth}
\centering\includegraphics[width=\linewidth, height=30pt]{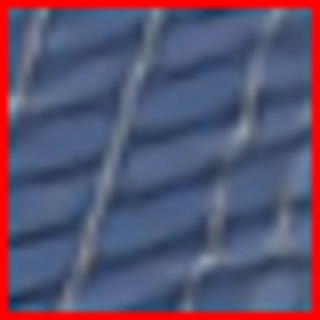}
\caption*{MaxSR-light\\\textcolor{red}{24.58}/\textcolor{red}{0.7620}}
\end{subfigure}
\end{minipage}
\caption{
Visual Results of Compared Methods for Lightweight Image SR.
}
\label{fig:visual_comparison_lightweight}
\end{figure*}
\section{Experiments}
\subsection{Datasets and Evaluation Metrics}
We use DIV2K \cite{Agustsson2017NTIRE2C}
and Flickr2K 
as training set, five standard benchmark datasets, Set5
, Set14
, BSD100
, Urban100 
and Manga109
, for performance evaluation
.
For classical image SR, we train our models using DIV2K or DIV2K plus Flickr2K following the different settings in literature.
For lightweight image SR, we train our models using DIV2K plus Flickr2K following literature \cite{DBLP:conf/eccv/ZhaoKHQD20,DBLP:conf/nips/LiZQJLJ20}.
HR images and corresponding LR images which are downsampled using bicubic interpolation are provided in DIV2K and Flickr2K dataset. Two kinds of data augmentation during training are used: 1) Rotate the images of 90, 180, 270 degrees; 2) Flip the images vertically. Peak signal-to-noise ratio (PSNR) and structural similarity index (SSIM) are used as evaluation metrics. The networks are trained using all three RGB channels and tested from Y channel of YCbCr color space of images.
\subsection{Implementation Details}
The MaxSR for classical image SR in experiments has (\(B=16\)) adaptive MaxViT blocks, which are divided into ($S=4$) stages, each stage has ($L=4$) adaptive MaxViT blocks. The width of network is set to (\(W=128\)), except the last \(3\times3\) convolutional layer outputs reconstructed HR color image.
The number of attention heads in adaptive block attention and adaptive grid attention is set to ($H=4$).
All layers use \(3\times3\) convolutional kernels except \(1\times1\) kernels in adaptive MaxViT blocks and HFFB.
The MaxSR-light for lightweight image SR in experiments reduces adaptive MaxViT blocks to (\(B=8\)), the number of adaptive MaxViT blocks in each stage to ($L=2$), and the width of network to (\(W=48\)). The other setting of MaxSR-light is the same as MaxSR.
Mini-batch size is set to \(32\) following the setting in SwinIR \cite{DBLP:conf/iccvw/LiangCSZGT21}. We use \(64\times64\) LR image patches and corresponding HR image patches for \(2\times\), \(3\times\), \(4\times\) and \(8\times\) scale factors to train our models. At each iteration of training MaxSR and MaxSR-light for each scale factor, we first sample an image pair from all pairs of corresponding HR and LR images uniformly, then sample a training patch pair from the sampled image pair uniformly.
Adam optimizer \cite{DBLP:journals/corr/KingmaB14} is used to train the network weights.
For $2\times$ scale factor, the learning rate is initialized to \(2\times10^{-4}\), and the $\times2$ model is trained for
\(500K\) or \(1500K\)
iterations when using DIV2K or DIV2K plus Flickr2K respectively,
the learning rate is step-decayed by \(2\)
after
[$250K, 400K, 450K, 475K, 500K$]
iterations or
[$750K, 1200K, 1350K, 1425K, 1500K$]
when using DIV2K or DIV2K plus Flickr2K for training respectively.
For $3\times$, $4\times$, $8\times$ scale factors, the models is finetuned from $2\times$ model, and the learning rate, the training iterations and the decay iterations are halved.
We implement our MaxSR with PyTorch framework.

\subsection{Ablation Study}
{\label{sec:ablation_study}}
We use models trained using DIV2K for classical image SR to do ablation study.
\subsubsection{Effect of Adaptive Block Attention or Adaptive Grid Attention}
As shown in Table \ref{tab:ablation_study_block_or_grid}, if we replace adaptive block attention by adaptive grid attention in MaxSR and denote the model as Ada-Gird-Attention
or replace adaptive grid attention by adaptive block attention in MaxSR and denote the model as Ada-Block-Attention,
the performance of models will degrade, which demonstrates adaptive block attention and adaptive grid attention are all indispensible to achieve good performance for SISR.
\subsubsection{Effect of Adaptiveness of MaxViT Block and Relative Position Embedding}
In this subsection, the models are trained using a big patch size of $128\times128$ to differentiate adaptive MaxViT block and original fixed-size MaxViT block during training, training batch size is reduced to $8$ correspondingly to maintain the same amount of data used in training.
When we adopt the strategy of original fixed-size MaxViT block, the fixed attention footage is set to $8\times8$.

The model trained and tested using original MaxViT block
is denoted as BP-Fix-SA-Tr-Fix-SA-Te.
The model trained and tested using proposed adaptive MaxViT block is
denoted as BP-Ada-SA-Tr-Ada-SA-Te, i.e., BP-MaxSR.
As shown in Table \ref{tab:ablation_study_adaptiveness_and_relative_position_embedding}, BP-MaxSR has achieved better performance than BP-Fix-SA-Tr-Fix-SA-Te across all datasets,
which demonstrates the effectiveness of adaptive Max-ViT block compared to original fixed-footage Max-ViT block \cite{DBLP:conf/eccv/TuTZYMBL22} for SISR.
We also train a model with original fixed-footage MaxViT block but test it with adaptive MaxViT block
and denote the model as BP-Fix-SA-Tr-Ada-SA-Te,
the performance of BP-Fix-SA-Tr-Ada-SA-Te will increase compared to that of BP-Fix-SA-Tr-Fix-SA-Te demonstrating the effectiveness of adaptiveness of MaxViT block used only in testing.

As shown in Table \ref{tab:ablation_study_adaptiveness_and_relative_position_embedding}, when we add relative position embedding into adaptive MaxViT block of BP-MaxSR and denote the model as BP-MaxSR-RPE,
which will achieve better performance than BP-MaxSR across all datasets. Although utilizing relative position embedding in adaptive MaxViT block will further boost performance for SISR,
we don't use it for MaxSR in this paper to trade performance for efficiency.

\subsection{Results on Classical Image SR}
We compare our method with
EDSR \cite{lim2017enhanced},
RDN \cite{zhang2018residual}, D-DBPN \cite{haris2018deep}, RCAN \cite{zhang2018image},
RRDB \cite{DBLP:conf/eccv/WangYWGLDQL18},
NLRN \cite{DBLP:conf/nips/LiuWFLH18},
RNAN \cite{DBLP:conf/iclr/ZhangLLZF19},
SAN \cite{DBLP:conf/cvpr/DaiCZXZ19},
OISR-RK3 \cite{DBLP:conf/cvpr/HeMWLY019},
IGNN \cite{DBLP:conf/nips/ZhouZZL20},
HAN \cite{niu2020single},
NLSN \cite{mei2021image},
IPT \cite{DBLP:conf/cvpr/Chen000DLMX0021},
SwinIR \cite{DBLP:conf/iccvw/LiangCSZGT21},
ENLCN \cite{DBLP:conf/aaai/XiaHTYLZ22},
ELAN \cite{DBLP:conf/eccv/ZhangZGZ22} 
to validate the effectiveness of MaxSR for classical image SR. We also adopt self-ensemble strategy \cite{lim2017enhanced} to improve MaxSR (MaxSR+) following the literature
.

The peak signal-to-noise ratio (PSNR) and the structural similarity index measure (SSIM) evaluation metrics on five benchmark datasets are shown in Table \ref{tab:classical_pnsr_ssim_table}.
MaxSR has established new state-of-the-art performance on different combinations of scale factors and benchmark datasets.

Figure \ref{fig:visual_comparison_classical} shows visual comparisons. As shown in
image \(babara\) from Set14,
image \(img092\) from Urban100,
our method reconstructs clearer edges, more textures, less artifacts, less blurring and ring effects of images than other methods.

\subsection{Results on Lightweight Image SR}
We compare our method with
DRRN \cite{DBLP:conf/cvpr/TaiY017},
CARN \cite{DBLP:conf/eccv/AhnKS18},
IDN \cite{hui2018fast},
IMDN \cite{2019Lightweight},
FALSR-A \cite{DBLP:conf/icpr/Chu0MXL20},
FALSR-C \cite{DBLP:conf/icpr/Chu0MXL20},
AAF-SD \cite{DBLP:conf/accv/WangWZYFC20},
MAFFSRN \cite{DBLP:conf/eccv/MuqeetHYKKB20},
PAN \cite{DBLP:conf/eccv/ZhaoKHQD20},
LAPAR-A \cite{DBLP:conf/nips/LiZQJLJ20},
RFDN \cite{DBLP:conf/eccv/LiuTW20},
AAF-L \cite{DBLP:conf/accv/WangWZYFC20},
A-CubeNet \cite{DBLP:conf/mm/HangLYC020},
LatticeNet \cite{DBLP:conf/eccv/LuoXZQLF20},
SwinIR-light \cite{DBLP:conf/iccvw/LiangCSZGT21},
ELAN-light \cite{DBLP:conf/eccv/ZhangZGZ22} 
to validate the effectiveness of MaxSR-light for lightweight image SR.

The performance metrics on five benchmark datasets are shown in Table \ref{tab:lightweight_pnsr_ssim_table}.
MaxSR-light has established new state-of-the-art performance
on different combinations of scale factors and benchmark datasets.

Figure \ref{fig:visual_comparison_lightweight} shows our Max-light achieves more better visual results compared to other methods.

\section{Conclusion}

In this paper, we propose a novel single image super-resolution model named as MaxSR based on recent hybrid vision transformer of MaxViT to utilize powerful representation capacity and self-similarity modelling ability of transformer models to boost performance for single image super-resolution.
In order to achieve better long-range dependency modelling ability/self-similarity modelling ability, we further improve MaxViT block to adaptive MaxViT block
which can integrate information from all windows
for each grid
and integrate information from all grids
for each window
to improve information propagation and global modelling of self-similarity
in an efficient way
which is helpful to performance
for SISR.
Experiments demonstrate the proposed model establish new state-of-the-art performance for classical image super-resolution (MaxSR) and lightweight image super-resolution (MaxSR-light) efficiently
even without adopting relative position embedding which can further increase performance for SISR.

\clearpage
\bibliographystyle{ACM-Reference-Format}
\bibliography{egbib}

\end{document}